\pdfoutput=1
\documentclass[11pt]{article}

\usepackage{setspace}
\PassOptionsToPackage{hyphens}{url}

\usepackage{assets/acl}

\usepackage{times}
\usepackage{latexsym}

\usepackage[T1]{fontenc}

\usepackage[utf8]{inputenc}

\usepackage{inconsolata}

\usepackage{microtype}

\usepackage{amsfonts}
\usepackage{amsthm}
\usepackage{amssymb}
\usepackage{amsmath}

\usepackage{tcolorbox}
\newcommand{\code}[1]{%
  {\small\texttt{#1}}%
 }

\usepackage{cleveref}
\crefname{section}{\S}{\S\S}
\Crefname{section}{\S}{\S\S}
\crefname{table}{Table}{}
\crefname{figure}{Figure}{Figs.}
\crefname{algorithm}{Algorithm}{}
\crefname{equation}{Eq.}{}
\crefname{appendix}{Appendix}{}
\crefname{thm}{Theorem}{}
\crefname{prop}{Proposition}{}
\crefname{cor}{Corollary}{}
\crefname{observation}{Observation}{}
\crefname{assumption}{Assumption}{}
\crefformat{section}{\S#2#1#3}
\crefformat{footnote}{#2\footnotemark[#1]#3}

\usepackage{multirow}
\usepackage{multicol}
\usepackage{adjustbox}

\usepackage{booktabs}
\usepackage{subcaption}
\usepackage{graphicx}
\usepackage{graphbox}

\usepackage{tabularray}
\usepackage{cellspace}
\newcommand\cincludegraphics[2][]{\raisebox{-0.3\height}{\includegraphics[#1]{#2}}}
\usepackage{setspace}
\usepackage{array}
\usepackage{titlesec}
\titlespacing{\paragraph}{%
    0pt}{%              left margin
    0.5\baselineskip}{% space before (vertical)
    0.5em}%               space after (horizontal)

\usepackage{enumitem}
\setenumerate{itemsep=\baselineskip, topsep=\baselineskip}

\usepackage{fnpct}

\usepackage{csquotes}  

\usepackage{todonotes}

\usepackage{siunitx}
\usepackage{xspace}
\DeclareSIUnit[quantity-product = {}, reset-math-version = false]\thousands{k}
\DeclareSIUnit[quantity-product = {}, reset-math-version = false]\x{x}
\DeclareSIUnit[quantity-product = {}, reset-math-version = false]\percent{\%}
\DeclareSIUnit[quantity-product = {}, reset-math-version = false]\million{M}
\DeclareSIUnit[quantity-product = {}, reset-math-version = false]\billion{B}

\newcommand{\integer}[1]{%
    \num[
        mode = text,
        round-mode=places,
        round-precision=0,
        group-separator={,},
        group-minimum-digits=4,
        ]{#1}\xspace
}
\newcommand{\float}[2][1]{\num[group-digits=false, round-precision=#1, round-mode=places]{#2}\xspace}

\xspaceaddexceptions{\textsubscript}

\def\name{\emph{Diable}\xspace}

\newcounter{myenumi}
\newcounter{myenumii}[myenumi]

\makeatletter
    \patchcmd{\HyField@FlagsRadioButton}{\HyField@SetFlag{Ff}{Radio}}{}{}{}
\makeatother

\title{\emph{Diable}: Efficient \underline{Dia}logue State Tracking as Operations on Ta\underline{ble}s}

\author{Pietro Lesci$^1$\thanks{{\hspace{0.15cm}}Work conducted during an internship at AWS AI Labs.} \quad Yoshinari Fujinuma$^2$ \quad Momchil Hardalov$^2$ \\ \bf Chao Shang$^2$ \quad Yassine Benajiba$^2$ \quad Llu\'{i}s M\`{a}rquez$^2$\\
  $^1$University of Cambridge \quad  $^2$AWS AI Labs\\
  \texttt{pl487@cam.ac.uk}; \\\texttt{\{fujinuy,momchilh,chshang,benajiy,lluismv\}@amazon.com}
\\  }

\begin{document}

\maketitle

\begin{abstract}
Sequence-to-sequence state-of-the-art systems for dialogue state tracking (DST) use the full dialogue history as input, represent the current state as a list with all the slots, and generate the entire state from scratch at each dialogue turn. This approach is inefficient, especially when the number of slots is large and the conversation is long. We propose \name, a new task formalisation that simplifies the design and implementation of efficient DST systems and allows one to easily plug and play large language models. We represent the dialogue state as a table and formalise DST as a table manipulation task. At each turn, the system updates the previous state by generating table operations based on the dialogue context. Extensive experimentation on the MultiWoz datasets demonstrates that \name \emph{(i)}~outperforms strong efficient DST baselines, \emph{(ii)}~is \qty{2.4}{\x} more time efficient than current state-of-the-art methods while retaining competitive Joint Goal Accuracy, and \emph{(iii)}~is robust to noisy data annotations due to the table operations approach.
\end{abstract}

% Github link
\noindent
\begin{tblr}{
  colspec = {Q[c,m]|X[l,m]},
  stretch = 0,
  %hlines = {red5, 1pt},
  %vlines = {red5, 1pt},
}%
 \cincludegraphics[width=1.2em, keepaspectratio]{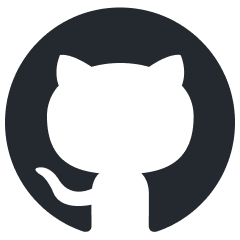} 
 & \setstretch{.5}\href{https://github.com/amazon-science/efficient-dialogue-state-tracking-by-sequential-information-processing}{\footnotesize\url{efficient-dialogue-state-tracking-by-sequential-information-processing}}\\
\end{tblr}

\section{Introduction}

\begin{figure}[!t]
    \centering
    \includegraphics[width=0.8\columnwidth]{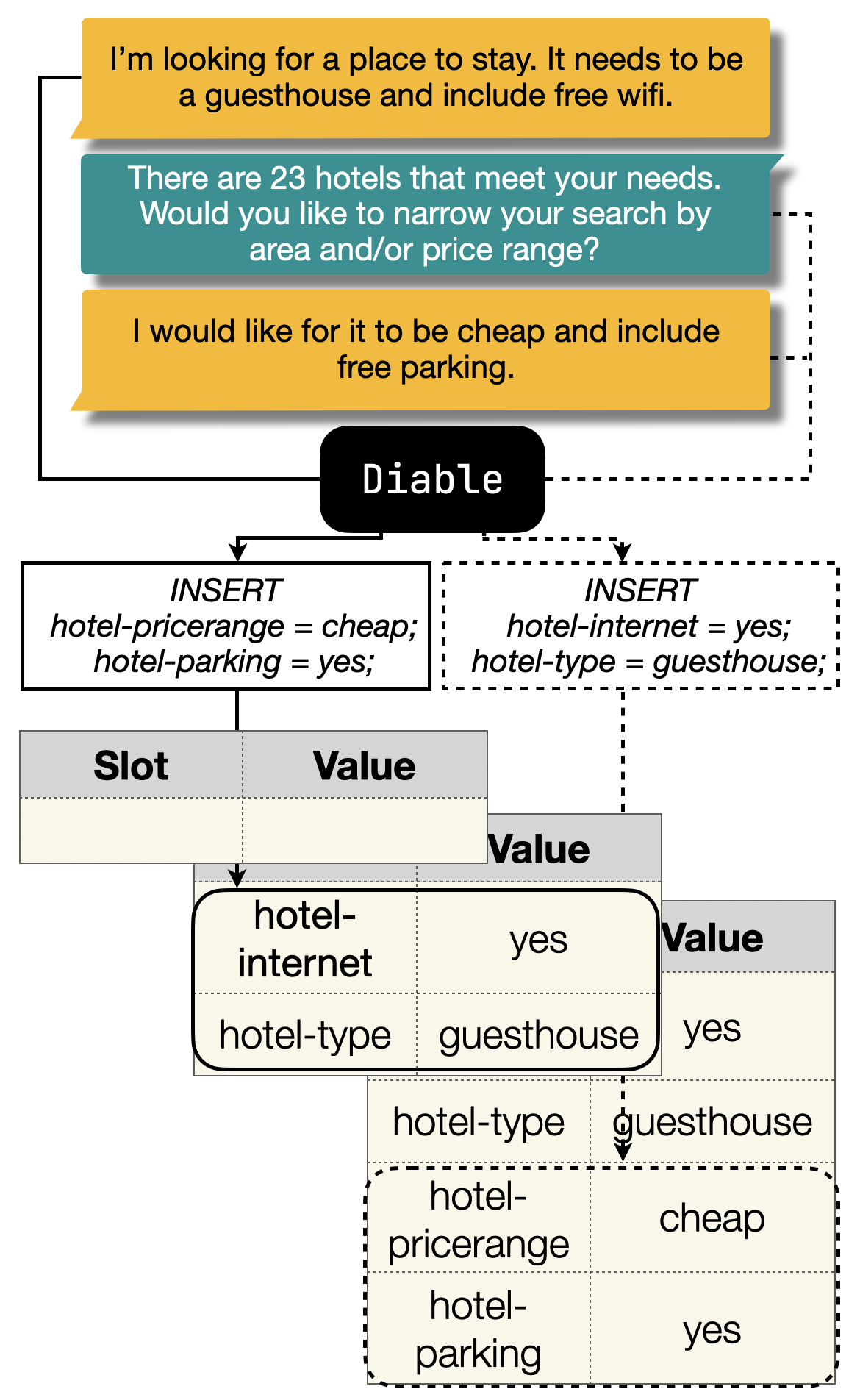}
    \caption{%
    \emph{Diable} approach to DST. The figure presents the first two turns of a dialogue (user's utterances are orange, system's are green). When the conversation starts, the state table is empty. At each dialogue turn, the system outputs a table update operation (either \code{INSERT} or \code{DELETE}), and the state is modified accordingly.    }
    %\vspace*{-.5em}
    \label{fig:diable_repr}
\end{figure}

\begin{figure}[!t]
    \centering
    \includegraphics[width=0.9\columnwidth]{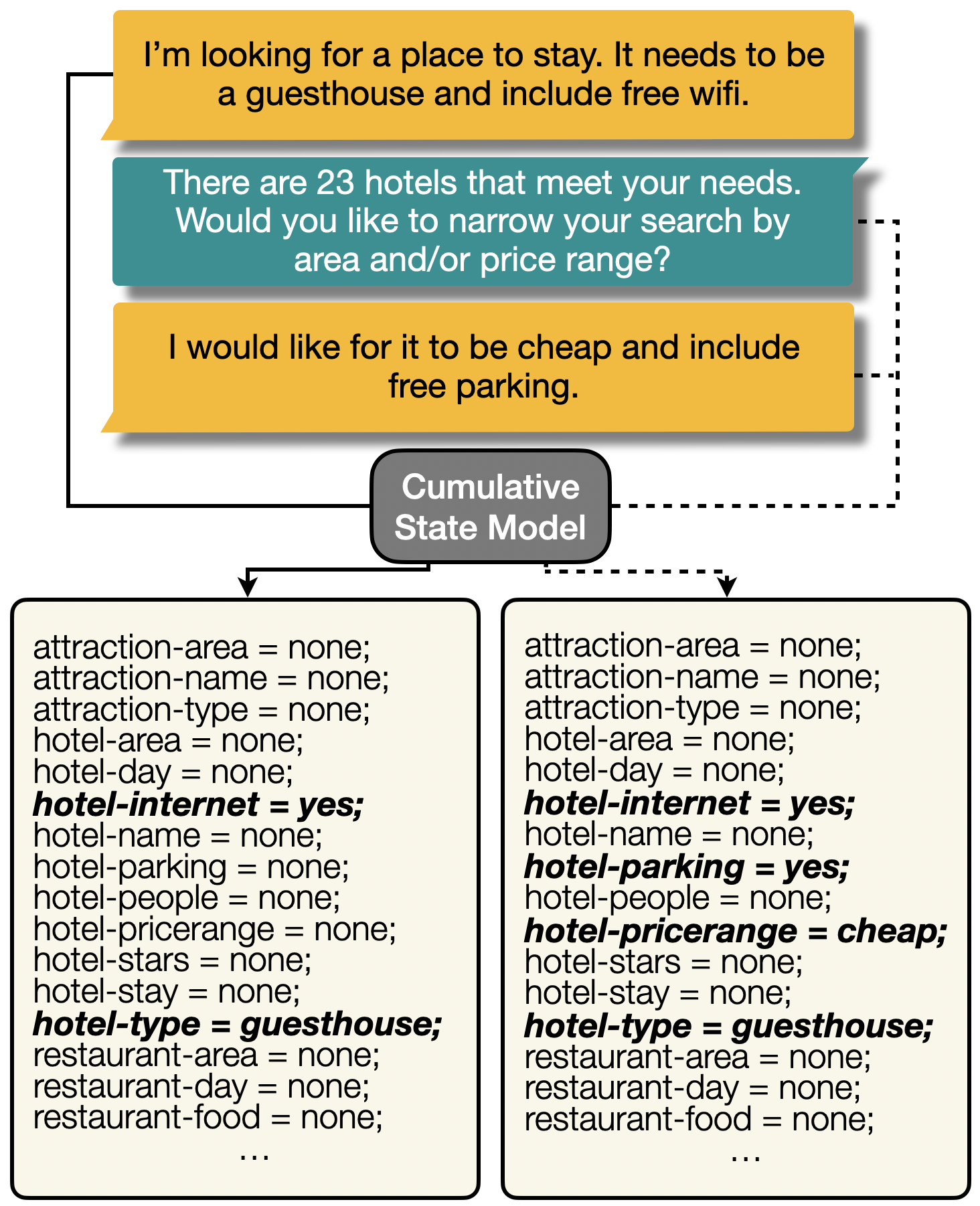}
    \caption{Cumulative state approach to DST. At each dialogue turn, the system outputs all the slots. Inactive slots are filled with a placeholder value (\code{none}).}
    %\vspace*{-.5em}
    \label{fig:cum_state}
\end{figure}

Dialogue state tracking~\citep[DST;][]{survey} is the task of tracking user requests from the dialogue history in the form of slot-value pairs~\citep{henderson-etal-2014-word, mrksic-etal-2015-multi, sgd-task-2020}. The slots are defined in a domain-specific schema and represent the fields that need to be extracted from the dialogue to execute queries in the backend and generate responses. Recent generative approaches to DST based on language models~\citep{wu-etal-2019-transferable, kim-etal-2020-efficient} often use the entire dialogue history as input and represent the state, at each turn, as the concatenation of all the slots in the schema, where inactive slots are reported with a placeholder value (see \Cref{fig:cum_state}). This representation is known as \emph{cumulative state}~\citep{hosseini-als-etal-2020-simple, feng-etal-2021-sequence, zhao-etal-2022-description} and implies the generation of all the states from scratch at each dialogue turn. This approach is computationally inefficient, especially for long conversations and large schemas.

We propose Efficient \underline{Dia}logue State tracking as Operations on Ta\underline{ble}s (\name, shown in \Cref{fig:diable_repr}), a novel task formulation and a new DST approach that better uses the generative capabilities of language models. Our approach simplifies the design and implementation of DST systems and works with any sequence-to-sequence model. Our intuition is that a DST system translates conversations into filters for database searches. Inspired by formal languages for databases and the recent success in applying sequence-to-sequence models to text-to-SQL tasks~\citep{yin-etal-2020-tabert, scholak-etal-2021-picard}, we represent the dialogue state as an implicit table and frame DST as a table manipulation task. At each turn, the system updates the previous state by generating update operations expressed in a simplified formal language based on the current dialogue context (see \Cref{fig:diable_repr}). \name is the first end-to-end DST system that outputs state operations and values jointly while processing all slots simultaneously. 

Based on extensive experimentation using the MultiWoz benchmark~\citep{budzianowski-etal-2018-multiwoz}, we show that \name can successfully and efficiently translate conversations into filters for database searches. Our approach minimises the number of input and output tokens required resulting in a significant efficiency gain (\qty{2.4}{\x} reduction in inference time compared to state-of-the-art cumulative state systems).

Our main contributions are as follows:
\begin{itemize}[itemsep=.05em]
    \vspace*{-.3em}
    \item We introduce a novel DST task formulation and a new system, \name,  specifically designed to enhance efficiency and leverage the capabilities of state-of-the-art sequence-to-sequence models.
    \item We show that our DST task formulation does not require ad-hoc data preprocessing, the full history, or extra supervision and works with any sequence-to-sequence model without requiring any architectural modification.
    \item We demonstrate that \name achieves better Joint Goal Accuracy on MultiWoz than other efficient baselines while being competitive with state-of-the-art cumulative state systems.
    \item We show that \name is robust to noise in the training data, resulting in more stable results across three versions of MultiWoz.
\end{itemize}

\section{A Taxonomy of DST Approaches}

The goal of DST systems is to handle long, diverse conversations in multiple domains with large schemas and unrestricted vocabulary, potentially without extra supervision~\citep{eric-etal-2020-multiwoz, sgd}. Achieving this goal has prompted the development of different DST approaches.

\paragraph{\emph{Ontology-based approaches}} \emph{treat DST as either a classification or a token classification task. They assume that all possible slot-value pairs are restricted to a fixed set, or ontology, either predefined or extracted from the training data. Classification-based approaches output a probability distribution over values given the dialogue context and a slot ~\citep{henderson-etal-2014-word} while token classification approaches output a probability distribution over slots for each token~\citep{liao-etal-2021-dialogue}}. 

\noindent The ontology-based formulation simplifies the DST task considerably, thus the performance of these systems is usually relatively high for specific datasets~\citep{zhong-etal-2018-global, ye-etal-2021-slot, ye-etal-2022-assist}. Complex dialogues with large schemas pose a significant challenge for traditional ontology-based approaches as they do not easily generalise to new domains nor scale to large ontologies~\citep{mrksic-etal-2017-neural, Rastogi2017, zhong-etal-2018-global, ye-etal-2021-slot}. For this reason, ontology-based approaches are out of scope for our paper.

\paragraph{\emph{Open-vocabulary approaches}} \emph{address these limitations by formulating DST as either a reading comprehension task wherein for each slot a span is extracted from the dialogue context \citep{gao-etal-2019-dialog, chao2019bert}, or as a generation task wherein a value for each slot is generated based on the dialogue history~\citep{wu-etal-2019-transferable}.}

\noindent By leveraging sequence-to-sequence models~\citep{brown_language_2020, lewis-etal-2020-bart, raffel-etal-2020-t5}, generative approaches have recently achieved states-of-the-art results~\citep{xu-hu-2018-end, lee-etal-2019-sumbt, chao2019bert, gao-etal-2019-dialog, wu-etal-2019-transferable, Kumar2020, heck-etal-2020-trippy, hosseini-als-etal-2020-simple, lee-etal-2021-dialogue, zhao-etal-2021-effective-sequence, zhao-etal-2022-description}. However, these methods predict the dialogue state from scratch at each turn and generate a value for each slot, even when a slot is not active (\Cref{fig:cum_state}). We argue (\Cref{sec:discussion}) that these are the main sources of inefficiencies of current DST systems. We compare \name with these methods in the \enquote{Cumulative State Models} section of \Cref{tab:results}.

\paragraph{\emph{Efficient approaches}} \emph{seek efficiency by minimising the number of values to generate, thus decomposing DST into two successive sub-tasks: state operation prediction and value generation. In this way, only the slots that need to be changed are considered for value generation~\citep{kim-etal-2020-efficient}.}

\noindent These approaches are the most related to \name in that they target efficiency. We compare them against \name in section \enquote{Efficient Models} of \Cref{tab:results}. Often, these methods~\citep{ren-etal-2019-scalable, zhu-etal-2020-efficient} use the cumulative state representation which is the primary source of inefficiencies (we discuss this issue in the context of \Cref{sec:results}, \Cref{tab:runtime}) and need to output operations for all slots.

For example,~\citet{kim-etal-2020-efficient} and~\citet{zeng-and-nie-2020} predict an operation for each slot in the input by adding a classification head on top of the tokens representing the individual slots and predict four kinds of state operations: \enquote{carryover}, \enquote{delete}, \enquote{dontcare}, and \enquote{update}. For those slots categorised as \enquote{update}, the contextual representation is further processed to decode the slot value. However, this approach limits the ability of such systems to deal with large schemas because the full schema needs to fit in the input context. Differently from these approaches, we remove the two-component structure by adopting a sequence-to-sequence approach that allows us to jointly generate operations and values for all slots simultaneously and works with any sequence-to-sequence model. Importantly, we only need to predict operations for the active slots (i.e., the slots actually mentioned in the conversation). 

\citet{lin-etal-2020-mintl} seek efficiency differently by introducing the notion of \enquote{Levenshtein belief span}. Based on the concept of \emph{belief span}~\citep{lei-etal-2018-sequicity} that reformats the dialogue state into a text span allowing models to generate slot values dynamically. They propose to only focus on the \emph{differences} between states at subsequent turns. We take this approach a step further by explicitly outputting operations for all slots changing from one turn to another simultaneously while retaining our minimal state representation.

\section{\name: Dialogue State Tracking as Operations on Tables}\label{sec:methodology}

We introduce a novel efficient formulation of the DST task and a system, \name, specifically designed to enhance efficiency and optimise the capabilities of state-of-the-art sequence-to-sequence models. In this section, we describe our approach, formalise the DST problem, and introduce the concepts of \emph{state as a table} and \emph{state operations}.

\subsection{Problem Definition}

The goal of DST is defined as learning a mapping from a dialogue context to a dialogue state. Specifically, let $D_{1:T} = (D_1, \dots, D_T)$ denote a dialogue of $T$ turns, where $D_t = (U_t, R_t)$ represent an utterance composed of the user query and system response at turn $t$, respectively. At turn $t$, the dialogue context, $C_t$,  is defined as the set of all the information available up to that turn. It always includes the current dialogue utterance, but can additionally contain the previous state, utterances from the dialogue history, and extra supervision (e.g., slot descriptions and the schema). We consider a dialogue context composed by only the previous dialogue turn(s) and the previous state, that is $C_t = \left(D_t, \mathcal{B}_{t-1}\right)$. We do not use any schema information and let the model learn it during training\footnote{Our preliminary study showed that passing the schema to the model has little effect on performance but hurts the model's efficiency as it needs to encode more tokens.}. The dialogue state at turn $t$ is defined as a set $\mathcal{B}_t = \{(s, v_t)|s \in \mathcal{S}_t\}$, where $\mathcal{S}_t \subseteq \mathcal{S}$ denotes the subset of active slots at that turn out of all the predefined slots in the schema and $v_t$ is the value corresponding to slot $s$.

\subsection{The \name Approach}\label{sec:approach}

In our approach, instead of directly outputting the dialogue state $\mathcal{B}$, we learn a mapping from the dialogue context, $C$, to a set of operations $\mathcal{O}$. At the beginning of each conversation the state table, $\mathcal{B}_0$, is initialised empty. At turn $t$, based on the dialogue context, $C_t$, the DST system generates the set of operations, $\mathcal{O}_t = \{O_1, \dots, O_{N_t}\}$, where $N_t$ is the number of slots that change between turn $t-1$ and $t$. Finally, the generated operations are applied to the previous state to get the new state. The tracker is expected to carry over the previously extracted slots into the current state, i.e.,~the state at each turn includes all the slots active since the beginning of the conversation up to that point. 

We operationalise this process by framing it as a sequence-to-sequence task in which a model, $f_\theta$, receives a textual representation of the dialogue context, $\tau_c(C_t)$, and outputs a textual representation of the operations needed, $\tau_s(\mathcal{O}_t)$, where $\tau_c$ and $\tau_s$ are the templating functions that convert the dialogue context and state operations to a string, respectively. We provide more details about these functions in \Cref{appendix:data_preprocessing}. The structure of the system can be described as follows
\begin{align}
    \tau_s(\mathcal{O}_t) &= f_\theta(\tau_c(C_t)) \\
    \mathcal{B}_t &= \text{Interpreter}(\tau_s(\mathcal{O}_t), \mathcal{B}_{t-1})\label{eq:state}
\end{align}
where in \cref{eq:state} we use an operation interpreter to parse the string representation of the operations and apply them to the previous state. Based on the definition of the state operations, the operation interpreter can be based on different formal languages (e.g., Regular Expressions, SQL). 

We use T5v1.1~\citep{raffel-etal-2020-t5} as the backbone for \name. During training, we use teacher forcing~\citep{goyal_professor_2016} and pass the oracle dialogue state in the input, $\mathcal{B}_{t-1}$. At test time, we pass the previously predicted state, $\hat{\mathcal{B}}_{t-1}$, instead. To learn the model, we optimise the negative log-likelihood of the state operations given the dialogue context, that is
\begin{equation}
    \mathcal{L}(\theta)_t = -\log P(\tau_s(\mathcal{O}_t)|\tau_c(C_t))
\end{equation}
where $f_\theta$ is used to parameterise the probabilistic model $P$. We use the Adafactor optimiser~\citep{pmlr-v80-shazeer18a} with no weight decay and we set the learning rate to $10^{-4}$ with a constant schedule. We fix the training budget at \qty{40}{\thousands} optimiser steps and set the batch size to \integer{32}. We generate the output sequence using beam search decoding with \integer{4} beams. We describe in detail the training and inference processes in \Cref{appendix:training_details}.

\subsection{Representing the State as a Table} 

In our approach, we represent the dialogue state as a table that is sequentially updated. Specifically, a state, $\mathcal{B}$, is represented by a simple two-column table in which the first column is used for the slot name and the second for the slot value (\Cref{fig:diable_repr}). We define the slot name as the concatenation of domain and slot separated by a dash, e.g.,~\code{restaurant-area} (see \Cref{appendix:data_preprocessing}). 
The state table is passed into the dialogue context by simple \enquote{linearisation}~\citep{suhr-etal-2020-exploring, scholak-etal-2021-picard, shaw-etal-2021-compositional}: the rows are converted to slot-value tuples, cast to a string using the template \code{\{slot\} = \{value\}}, and concatenated together using \code{;} as a separator\footnote{More complex table encoding methods can be applied~\citep{herzig-etal-2020-tapas,yang-etal-2022-tableformer,nassar-etal-2022-tableformer}.  See the discussion in \Cref{sec:future_directions}.}. During the linearisation, we randomise the order of the rows to avoid overfitting to specific positional biases.

\subsection{State Tracking via Table Operations}

We introduced how we operationalise table operations as a combination of strings defining the operations and an interpreter that applies the operations to the state. Specifically, in our implementation of \name, we use a simplified formal language consisting of two slot-level operations, \code{INSERT} and \code{DELETE}, and a simple regex-based interpreter. The choice of the available operations is motivated by the nature of the MultiWoz datasets which include mostly insertions and deletions. We use the \code{INSERT} operation also to update the value of slots that are already present in the state. When no operation is needed, the target is defined as the literal string \code{none}.
Updates are less frequent and are caused mostly by inconsistent annotations. In our preliminary experiments, we empirically found that adding an \code{UPDATE} operation does not improve performance despite adding complexity; thus, we decided to not use it. We emphasise that the specific definition of the operations is not critical for the efficiency of our method and it can be easily adapted to any specific use case.
To convert operations to strings we use the template \code{\{command\} \{slot\} = \{value\}}. If multiple operations need to be applied, we concatenate them using \code{;} as a separator (see \Cref{appendix:data_preprocessing}). We define the target sequence as the concatenation of all the slot-level operations. Since the order in which the operations are applied does not affect the output, we randomise their position during training. 

\begin{table*}[!t]
    %\sffamily
    \centering
    %\begin{adjustbox}{width=\textwidth}
     \resizebox{\textwidth}{!}{%
    \setlength{\tabcolsep}{3pt}
\begin{tabular}{ll|>{\centering\arraybackslash}p{5cm}c|rrr}
\toprule
\textbf{Model} & \textbf{Architecture} & \textbf{Extra Supervision} & \textbf{Context} ($C_t$) & \textbf{2.1} & \textbf{2.2} & \textbf{2.4} \\
\midrule
\midrule
\multicolumn{2}{l}{\textbf{Cumulative State Models}} \\
\midrule
TRADE~{\small \citep{wu-etal-2019-transferable}} & BERT-base (110M) & Schema & $D_{1:t}$ & $^\ddagger45.60$ & $^\ddagger45.40$ & $^\ddagger55.10$ \\

SUMBT~{\small \citep{lee-etal-2019-sumbt}} & BERT-base (110M) & Schema & $D_{1:t}$ & $^\ddagger49.20$ & $^\ddagger49.70$ & $^\ddagger61.90$ \\

DS-DST~{\small \citep{zhang-etal-2020-find}}  & BERT-base (110M) & Schema + Pick. & $D_{1:t}$ & $^\ddagger51.21$ & $^\ddagger51.70$ & - \\

TripPy~{\small \citep{heck-etal-2020-trippy}} & BERT-base (110M) & - & $D_{1:t}$ & $^\ddagger55.30$ & $^\dagger53.52$ & $^\ddagger59.60$ \\

SAVN~{\small \citep{wang-etal-2020-slot}} & BERT-base (110M) & Schema & $D_{1:t}$ & $^\ddagger54.50$ & - & $^\ddagger60.10$ \\

Seq2Seq-DU~{\small \citep{feng-etal-2021-sequence}} & 2x BERT-base (220M) & Schema + Desc. & $D_{1:t}$ & $^\dagger56.10$ & $^\ddagger54.40$ & - \\

\midrule

SimpleTOD~{\small \citep{hosseini-als-etal-2020-simple}} & GPT-2 (117M) & Schema & $D_{1:t}$ & $^\ddagger50.3/55.7$ &  $^\star54.02$ & - \\

AG-DST~{\small \citep{tian-etal-2021-amendable}} & PLATO-2 (310M) & Schema & $D_{t-1:t}+\mathcal{B}_{t-1}$ & - & $57.26$ & - \\

\midrule

DaP (seq.)~{\small \citep{lee-etal-2021-dialogue}} & T5-base (220M) & Schema + Desc.  & $D_{1:t}$ & -  & $51.20$ & - \\

DaP (ind.)~{\small \citep{lee-etal-2021-dialogue}} & T5-base (220M) & Schema + Desc.  & $D_{1:t}$ & $56.66$  & $\underline{57.60}$ & - \\

Seq2seq~{\small \citep{zhao-etal-2021-effective-sequence}} & T5-base (220M) & Pre-training  & $D_{1:t}$ & $^\diamond52.80$  & $\underline{57.60}$ & $67.10$ \\

\textit{light}Cumulative (Our impl.) & T5v1.1-base (247M) & - &$D_{1:t}$ & $53.91_{\pm 0.63}$ & $57.01_{\pm 0.45}$ &  $67.56_{\pm 0.52}$ \\

D3ST~{\small \citep{zhao-etal-2022-description}} & T5-base (220M) & Schema + Pick. + Desc.  & $D_{1:t}$ & $54.20$ & $56.10$ & $\underline{72.10}$\\

%\midrule
%\multicolumn{2}{l}{\hspace{1mm} \textbf{+ scale-up model size}} \\
%\midrule
%D3ST~{\small \citep{zhao-etal-2022-description}} & T5-xxl (11B) & Schema + Pick. + Desc.  & $D_{1:t}$ & $\underline{57.80}$ & $\underline{58.70}$ & $\underline{75.90}$\\

\midrule
\midrule
\multicolumn{2}{l}{\textbf{Efficient Models}} \\
\midrule
MinTL~{\small \citep{lin-etal-2020-mintl}} & BART-large (406M) & - & $\mathcal{B}_{t-1}$ & $53.62$ & - & - \\

SOM-DST~{\small \citep{kim-etal-2020-efficient}} & BERT-base + GRU (113M) & Schema & $D_{t-4:t}+\mathcal{B}_{t-1}$ & $^\ddagger53.68$ & $^\star53.81$ & $^\ddagger66.80$\\

Transf.-DST~{\small \citep{zeng-and-nie-2020}}  & BERT-base (110M) & Schema & $D_{t-4:t}+\mathcal{B}_{t-1}$ & $\mathbf{55.35}$ & - & - \\
\midrule
\multirow{2}{*}{\textbf{Diable} (Ours)} & T5v1.1-base (247M) & - & $\mathcal{B}_{t-1}$ & $^\diamond53.91_{\pm 0.70}$ & $56.30_{\pm 0.67}$ &  $70.03_{\pm 0.95}$ \\

& T5v1.1-base (247M) & - & $D_{t-4:t} + \mathcal{B}_{t-1}$ & $^\diamond53.97_{\pm 0.66}$ & $\mathbf{56.48}_{\pm 0.57}$ &  $\mathbf{70.46}_{\pm 1.18}$ \\

\bottomrule
\end{tabular}
    %\end{adjustbox}
    }
    \caption{JGA on the test sets of MultiWoz (2.1, 2.2 and 2.4) for models trained on the respective training sets (note that 2.1 and 2.4 share the same training data). Baseline results reported from the original papers or, when not available, from $\star$: \citet{tian-etal-2021-amendable}, $\dagger$: \citet{wang-etal-2022-luna}, $\ddagger$: \citet{zhao-etal-2021-effective-sequence}. The column \enquote{Context} reports the dialogue context: $D$ and $\mathcal{B}$ denote the dialogue utterances and the set of previous states, respectively. The notation ${}_{i:j}$ indicates turns from $i$ to $j$ (included). The column \enquote{Extra supervision} reports the additional information used  (e.g., data augmentation, pre-training, etc.). For Multiwoz 2.1 most baselines use data preprocessing; we denote methods that do not use data preprocessing with $\diamond$. \underline{Underlined} the best results overall; \textbf{bold} the best results within efficient methods.
    }
    \label{tab:results}
    \vspace*{-.5em}
\end{table*}

\section{Experiments}

In this section, we present our experimental setup and provide details about the baselines approaches.

\subsection{Datasets} 

The MultiWoz dataset~\citep{budzianowski-etal-2018-multiwoz} is a collection of \qty{10}{\thousands} multi-domain task-oriented human-to-human conversations. It is one of the most used benchmarks in the DST literature~\citep{survey}. Nonetheless, it is known to contain annotation errors and previous work proposed different versions~\citep{eric-etal-2020-multiwoz, mwoz_23, ye-etal-2022-multiwoz} and data normalisation procedures\footnote{Most notably the \href{https://github.com/jasonwu0731/trade-dst/blob/master/utils/fix_label.py}{TRADE scripts} from~\citet{wu-etal-2019-transferable} to normalise both text and labels.} to mitigate this issue. Thus, it is difficult to have a fair comparison of results across the literature. 
Following the MultiWoz convention~\citep{wu-etal-2019-transferable}, we filter out dialogues in the \enquote{bus}, \enquote{police}, and \enquote{hospital} domains (and the respective slots from multi-domain dialogues), and remove the invalid dialogue \code{SNG01862.json}. We experiment with multiple versions (2.1, 2.2, and 2.4) and use the data as-is (see~\Cref{appendix:data_preprocessing}). To construct the training set, we extract the operations automatically from the dataset.

\subsection{Evaluation}\label{sec:evaluation}

We use Joint Goal Accuracy~\citep[JGA]{henderson-etal-2014-word} as the main metric for all experiments: it measures the proportion of turns for which the predicted state (slot-value pairs) exactly match the gold label. At each turn, for each slot, a list of acceptable values is included in the annotation (e.g.,~\code{hotel-name: ["marriott", "marriott hotel"]}). We consider a value correct if it exactly matches one of the available options. Importantly, we perform an uncased evaluation since the annotation casing is not consistent.

\subsection{Cumulative State and Efficient Baselines}

We compare our results with a set of strong cumulative state models (i.e.,~models that use all previous turns and output a value for each slot at each turn, see \Cref{fig:cum_state}), and efficient baseline models. We also implement our own version of a cumulative state model and its \enquote{lighter} variant, \emph{light}Cumulative: the state does not include the inactive slots. In all our experiments, the full cumulative models underperform \emph{light}Cumulative while being less efficient ($\sim$\qty{1.18}{\x} slower). Thus, we only report the results  of \emph{light}Cumulative, effectively selecting a stronger baseline.

In the upper part of \Cref{tab:results} (\enquote{Cumulative State Models}), we include results from state-of-the-art generative cumulative state models. In each section, we report details and results for encoder-based, sequence-to-sequence, and T5-based models, respectively. The latter class of models based on T5 is related to our implementation of \name in that they share the same backbone. However, they are not directly comparable due to the additional text preprocessing and label normalisation. The results of our own re-implementation of a cumulative state model, \emph{light}Cumulative, are directly comparable as we adopted the same experimental conditions.

In the bottom part of \Cref{tab:results} (\enquote{Efficient Models}), we report the JGA of the latest generative efficient DST approaches in the literature. Despite being related to our implementation of \name, these approaches are not directly comparable since they rely on additional information (e.g. the schema) or are based on a different backbone model.

\section{Results}\label{sec:results}

In this section, we discuss our experimental results. In \Cref{tab:results}, we summarise the JGA on three versions of MultiWoz (2.1, 2.2, and 2.4) for both \name and the baseline models. The results for the baseline models are taken from previous work~\citep{tian-etal-2021-amendable, wang-etal-2022-luna, zhao-etal-2021-effective-sequence} when better or if missing for a particular version in the original papers. The results for \name and \emph{light}Cumulative implementation are averaged across \integer{5} random seeds.

\subsection{\name and Cumulative State Models} 

We compare \name's performance to cumulative state models, i.e., models that have access to all previous turns in the conversation. We emphasise that \name uses none or a limited number of previous turns and thus has less context with respect to these models. On one hand, our goal is to evaluate the trade-off between efficiency and performance; on the other hand, to study the capability of the system to generate correct table operations.

The cumulative state model results are shown in the first part of \Cref{tab:results}. First, D3ST~\citep{zhao-etal-2022-description} achieves the best JGA score on MultiWoz 2.4 when the backbone is T5-base. Similarly to \name, D3ST is based on a T5 model; however, it has access to more information such as the schema, the slot descriptions, and the list of possible values (\enquote{picklist}) for categorical slots. Nonetheless, \name scores within \integer{1} standard deviation in terms of JGA, while being more than \qty{2}{\x} more efficient.

\begin{table}[!t]
 %\sffamily
 \centering
 \begin{adjustbox}{width=0.9\columnwidth}
    \begin{tabular}{lrr}
\toprule
\textbf{Context} & \textbf{Runtime (ms)} & \textbf{Speedup ($\uparrow$)} \\
\midrule
\midrule
\textbf{Cumulative}\\
\midrule
\hspace{3mm} $D_{1:t}$ & 40.83 & 1.00 \\
\midrule
\midrule
\textbf{\textit{light}Cumulative}\\
\midrule
\hspace{3mm} $D_{1:t} + \mathcal{B}_{t-1}$ & 34.28 & 1.19 \\
\hspace{3mm} $D_{1:t}$ & 33.65 & 1.21 \\
\hspace{3mm} $D_{t-4:t} + \mathcal{B}_{t-1}$ & 32.56 & 1.25 \\
\hspace{3mm} $\mathcal{B}_{t-1}$ & 30.62 & 1.33 \\
\midrule
\midrule
\textbf{\emph{Diable}}\\
\midrule
\hspace{3mm} $D_{1:t} + \mathcal{B}_{t-1}$ & 19.59 & 2.09 \\
\hspace{3mm} $D_{1:t}$ & 19.41 & 2.11 \\
\hspace{3mm} $D_{t-4:t} + \mathcal{B}_{t-1}$ & 18.29 & 2.23 \\
\hspace{3mm} $\mathcal{B}_{t-1}$ & 17.17 & 2.38\\
\bottomrule
    \end{tabular}
 \end{adjustbox}
 \caption{Median instance-level runtime in milliseconds and relative speed-up \textit{vs} a cumulative state baseline.
 }
 \label{tab:runtime}
 \vspace*{-.1em}
\end{table}

When the backbone model of D3ST~\citep{zhao-etal-2022-description} is T5-xxl (\qty{11}{\billion}), it scores \float[2]{57.80}, \float[2]{58.70}, and \float[2]{75.90}, respectively, on the three versions of the MultiWoz dataset. These scores are significantly higher than all other baselines. However, this improvement is solely due to increasing the model size, and we argue that the same performance improvement can be achieved by scaling the backbone of \name to larger models. In particular, error analysis shows that most of the errors in our instantiation of \name-based systems are due to the model not recognising active slots (\enquote{under-predictions}). A larger backbone model can alleviate this issue by picking up less obvious contextual cues. Finally, the difference is more significant for version 2.1 because D3ST also applies text preprocessing, as used in other baselines. Moreover, baselines that use smaller models (the first part of Table~\ref{tab:results}) consistently score lower than those based on the larger and better pre-trained T5 checkpoints. The only exception is AG-DST~\citep{tian-etal-2021-amendable} but their backbone model has \qty{310}{\million} parameters.

We further compare \name to our implementation of a cumulative state T5-based model (\emph{light}Cumulative). This comparison is fairer, as the models are put in the exact same experimental conditions. Our goal here is to quantify the improvements due to our proposed approach isolating additional effects from model pre-training and architectural changes. The results show that our \name approach has a significantly better JGA ($+$\integer{3} absolute points) on the less noisy version of MultiWoz (i.e., 2.4) and has similar performance on 2.1 and 2.2, while still being more efficient.

Next, we compare \name with two other strong models based on the T5 architecture (making them directly comparable, besides the preprocessing steps): \emph{DaP (ind)}~\citep{lee-etal-2021-dialogue} and \emph{Seq2seq}~\citep{zhao-etal-2021-effective-sequence}. Both models achieve a slightly higher JGA than \name on 2.2 (\integer{1} point absolute); however, they are again less efficient and have access to a larger context. \emph{DaP} relies on slot descriptions (thus, the schema) and runs inference once for every slot, which is not scalable to large schemas. The improvements in \emph{Seq2seq} are likely due to its extensive DST-specific pre-training.

Our results confirm that \name-based systems while being efficient, achieve competitive JGA on MultiWoz as compared to both other strong efficient DST baselines and cumulative state state-of-the-art DST systems without requiring any ad-hoc data preprocessing, access to the full history, extra supervision, or large backbone models.

\subsection{\name and Efficient Models} 

Comparing \name to other efficient state-of-the-art DST models, that are based on state operations, we see significant improvements up to almost \integer{4} JGA points on version 2.4 (shown in the  \enquote{Efficient Models} section of \Cref{tab:results}). Only Transformer-DST~\citep{zeng-and-nie-2020} is able to outperform our model on 2.1. However, they use data preprocessing (text and label normalisation) and extra supervision (schema). This model is an improved version of SOM-DST~\citep{kim-etal-2020-efficient}, therefore the same argument applies to the latter, which achieves slightly lower performance even using the same extra supervision and text normalisation.

\subsection{Latency Analysis} 

\Cref{tab:runtime} reports the median\footnote{We report the median as the distribution of the inference time is left-skewed.} inference time and the speed-up factor of \name relative to \emph{light}Cumulative. Our approach is more than \qty{2}{\x} faster, even when using the full history as context. These results clearly show that the biggest efficiency gains are obtained by shortening the output sequence, that is, replacing the cumulative state with state operations. Consequently, adding only the last $N$ turns comes at a small cost for a \name while potentially helping the model to recover values not present in the current dialogue context.  Using the Inference Time Complexity (ITC) notation introduced by~\citet{ren-etal-2019-scalable}, our proposed approach has $\Omega(1)$ and $O(N)$, where $N$ is the number of slots in the schema, as the best and worst case ITC, respectively. Whereas, SOM-DST and Transformer-DST have a best-case ITC of $\Omega(N)$.

\begin{table}[!t]
    \centering
    \small
    \begin{adjustbox}{width=0.9\columnwidth}
    \begin{tabular}{lrr}
\toprule
 \textbf{Train $\to$ Test} & \textbf{\textit{light}Cumulative} & \textbf{\textit{Diable}} \\
 \midrule
 \midrule
\textbf{Context: $D_{1:t}$}\\
\midrule
\hspace{3mm} $2.2 \to 2.2$ & $57.01_{\pm 0.45}$ & $55.63_{\pm 0.68}$ \\
\hspace{3mm} $2.2 \to 2.4$ & $63.11_{\pm 0.83}$ & $64.95_{\pm 0.55}$ \\
\midrule
\midrule
\textbf{Context: $\mathcal{B}_{t-1}$}\\
\midrule
\hspace{3mm} $2.2 \to 2.2$ & $56.50_{\pm 0.47}$ & $56.30_{\pm 0.67}$ \\
\hspace{3mm} $2.2 \to 2.4$ & $63.52_{\pm 0.96}$ & $66.13_{\pm 0.97}$ \\
\bottomrule
\end{tabular}
    \end{adjustbox}
    \caption{Effect on JGA (mean $\pm 1$ standard deviation) of different context and state representations. 
    %The experimental setting for all experiments is exactly the same.
    %When evaluated on the cleaned version 2.4, efficient models consistently match or outperform cumulative state models.
    }
    \label{tab:robustness}
    %\vspace*{-.5em}
\end{table}

\subsection{Robustness to Noisy Annotations} 

\Cref{tab:robustness} compares the performance of models trained on MultiWoz 2.2 with different context and state representations. Notably, when evaluated on the cleaner 2.4 version (bottom row for both parts of the table), \name consistently outperforms \emph{light}Cumulative. In fact, regardless of the dialogue context, \name achieves a better JGA on 2.4. We hypothesise that the lower accuracy of \emph{light}Cumulative is due to overfitting the noisy annotations of the training set. In particular, we think that since it generates the full state from scratch at every turn, the decoder might learn wrong correlations amongst slots that are wrongly annotated in the training set. For example, \code{hotel-type} and \code{attraction-type} are inconsistently and sparsely annotated in the training set, while in the test set of version 2.4 they tend to appear almost always together with the respective \code{hotel-name} and \code{attraction-name} slots. Thus, a cumulative state model can learn to not generate one when the other is present. Instead, being \name based on state \emph{changes}, we presume that it learns to treat slots more independently.

\section{Discussion}\label{sec:discussion}

Our task formalisation is intuitively simple and is especially beneficial for large pre-trained sequence-to-sequence models. First, the state is expanded sequentially and thus only includes the necessary slots. This minimises the size of the input context, allowing the models to scale to larger schemas before reaching their maximum input length.  Second, since the model needs to focus on the state \emph{changes}, the decoder only needs to generate operations for a limited number of slots (previous slots persist implicitly in the state, no need for explicit \enquote{carryover} operations). Third, our system is general in that it deals with span-based and categorical slots in the same way, and outputs both the operations and the slot-value pairs in a single forward pass, without the need for specialised architectures. Finally, since not all pre-defined slots are needed in the input, we do not have to access the schema beforehand, and thus it can be learned from the data directly.

\begin{table}[!t]
    \centering
    \small
    \begin{adjustbox}{width=0.9\columnwidth}
    \begin{tabular}{lc|rr}
\toprule
\textbf{Context} &  \textbf{Test Set} & \multicolumn{2}{c}{\textbf{JGA}} \\
% \cmidrule(lr){3-4}
%  &   &  \textbf{base} & \textbf{large} \\
\midrule
\midrule
\textbf{Cumulative state} &   &  \texttt{base} & \texttt{large} \\
\midrule
$D_{1:t}$ &     2.2  & $57.37$ & $57.01$ \\
$D_{1:t}$ &     2.4  & $65.82$ & $63.11$ \\
\midrule
\midrule
\textbf{\emph{Diable}} &   &  \texttt{base} & \texttt{large} \\
\midrule
$D_{t-4:t} + \mathcal{B}_{t-1}$ & 2.2  & $56.74$ & $56.48$ \\
$D_{t-4:t} + \mathcal{B}_{t-1}$ & 2.4  & $65.01$ & $65.35$ \\
\bottomrule
\end{tabular}
    \end{adjustbox}
    \caption{MultiWoz 2.2 and 2.4 test set JGA for T5v1.1 \code{base} and \code{large} trained on the MultiWoz 2.2. 
    %Scaling up model size does not improve the performance.
    }
    \label{tab:base_large_comparison}
    %\vspace*{-.5em}
\end{table}

\subsection{Impact of the Dialogue History} \Cref{tab:robustness} compares the effect of the context size for both \emph{light}Cumulative and \name trained on version 2.2. Comparing the results from the upper and bottom parts of the table, we see that using only the previous state barely changes the JGA of \emph{light}Cumulative but benefits \name. We hypothesise that being a cleaner and more compact representation of the conversation, the previous state introduces less noise than the full history. This is especially true in conversations for which the value of a slot is changed or removed throughout the conversation. However, completely removing the dialogue history reduces the ability of the model to recover values referenced at the beginning of the conversation. We hypothesise that this negative effect is not too evident because of the entity bias present in the MultiWoz dataset~\citep{qian-etal-2021-annotation} that allows the model to memorise and correctly predict values for certain slots even when not present in the dialogue context (\Cref{sec:performance_gap}). Finally, when evaluated on the cleaned version 2.4, \name consistently matches or outperforms \emph{light}Cumulative.

\subsection{Impact of the Model Size} \Cref{tab:base_large_comparison} compares the performance of the base and large version of T5v1.1 for both \emph{light}Cumulative and \name models. We find that scaling up model size does not improve JGA, however, we hypothesise that scaling it further can improve the performance similarly to D3ST~\citep{zhao-etal-2022-description}.

\subsection{Impact of the State Representation} When replacing the tabular state representation with a cumulative one in \name, \emph{ceteris paribus}, we find a \qty{3}{\percent} reduction in JGA for version 2.4 and up to \qty{5}{\percent} for other versions.  Specifically, at the beginning of the conversation, the state includes all the slots with the \code{none} value. In this case, the \code{INSERT} operation is unchanged while the \code{DELETE} operation becomes an update with a \code{none} value. 

\begin{table}[!t]
    \centering
    \small
    \begin{adjustbox}{width=0.9\columnwidth}
    \begin{tabular}{lrr}
\toprule
 \textbf{Train $\to$ Test} & \textbf{Predicted} & \textbf{Gold} \\
\midrule
$2.1 \to 2.1$ & $53.91_{\pm 0.70}$ & $80.65_{\pm 0.24}$ \\
$2.1 \to 2.4$ & $70.03_{\pm 0.95}$ & $90.14_{\pm 0.30}$ \\
$2.2 \to 2.2$ & $56.30_{\pm 0.67}$ & $82.50_{\pm 0.28}$ \\
$2.2 \to 2.4$ & $66.13_{\pm 0.97}$ & $88.29_{\pm 0.35}$ \\
\bottomrule
\end{tabular}
    \end{adjustbox}
    \caption{JGA (mean $\pm 1$ standard deviation) with gold and the predicted previous state in the input context. %The experimental setting kept exactly the same. 
    %There is the potential to improve by adopting recent methodologies targeted at reducing error propagation.
    }
    \label{tab:performance_gap}
    %\vspace*{-.5em}
\end{table}

\subsection{Error Propagation}\label{sec:performance_gap}

\name, like any recursive state model~\citep{zhao-etal-2021-effective-sequence}, is affected by error propagation: since we pass the previous predicted state at each turn, errors can be persisted. We measure the potential gains stemming from completely avoiding the error propagation by using the \emph{gold} previous state rather than the predicted one in the dialogue context. \Cref{tab:performance_gap} reports the upper bound on JGA for our simple \name instantiation and highlights that there is potential to improve JGA by adopting recent methodologies targeted at reducing error propagation~\citep{monet}.

In our experiments, we identify two main sources of error propagation that account for more than \qty{60}{\percent} of the total mistakes: state \enquote{under-prediction} (i.e., the model does not recognise that a certain state is active)  and value misprediction. Under-prediction happens when the system is unable to recognise that specific slots are active. Since MultiWoz presents a strong entity bias---e.g., \enquote{Cambridge} appears in \qty{50}{\percent} of the destination cities in the training data~\citep{qian-etal-2021-annotation}---a possible direction to address this issue is to use data augmentation methods targeted at reducing entity bias and annotation inconsistency~\citep{summerville-etal-2020-tame, lai-etal-2022-controllable} by improving the overall slot recall. Value misprediction happens when the value for a correctly predicted slot is wrong. This is especially evident when the same slot is discussed in multiple turns and its value can potentially change. One way to address this limitation is by automatically pre-selecting the previous dialogue turns to include the relevant information about a specific slot in the context window~\citep{yang-etal-2021-comprehensive, guo-etal-2022-beyond, wang-etal-2022-luna}. 

We do not constrain the generation in any way, and thus \name can generate invalid slots or values (e.g.,~\code{attraction-time}). In our experiments, errors due to invalid states are rare (less than 2\% of the total mistakes): in fact, using the schema to filter incorrectly predicted slots at each turn did not improve the JGA significantly (less than 1\%). There are several promising techniques that can further improve the performance of our system, at a minor efficiency cost, such as amendable generation~\citep{tian-etal-2021-amendable}, constrained generation~\citep{lin-etal-2020-commongen}, and schema descriptions~\citep{lee-etal-2021-dialogue, zhao-etal-2022-description}. Finally, with larger schemas and more diverse conversations, constraining the set of values that the model can predict can potentially further improve performance and safety.

\subsection{Future Directions}\label{sec:future_directions}

In \Cref{sec:results}, we showed that \name is an effective DST approach, at the same time it is competitive with budget-matched (in terms of parameter count) cumulative state baselines. We emphasise that our goal is not to reach state-of-the-art JGA on the MultiWoz dataset. We intentionally keep our \name-based models as simple as possible, by not adding extra supervision signals, to clearly measure the effectiveness of our approach. However, the benefits coming from \name can be easily added on top of other methods. We believe our approach can be improved and expanded in several ways.

\paragraph{Explicitly Modelling Slot Dependence.} \name treats slots independently of each other and implicitly uses the model's capability of learning their co-occurrence patterns. However, as the schema becomes larger and the dialogues longer, slot dependence becomes more complex and the model might fail to learn it effectively. Explicitly modelling the slot dependence can potentially improve performance, robustness (to spurious correlations), and efficiency. For example, selecting only relevant turns from the dialogue history as context to predict slot values. 
In our experiments, we show consistent improvement across all MultiWoz versions by adding the previous 4 dialogue turns in the dialogue context (\Cref{tab:results} -- last 2 rows). However, this simple heuristic might be suboptimal when the schema is large and the dialogue is long because relevant turns may not be the immediately preceding ones and we might add irrelevant context or omit relevant information. Instead, adopting a more granular turn selection method based on the slot dependence~\citep{yang-etal-2021-comprehensive, guo-etal-2022-beyond} can improve both performance and efficiency.

\paragraph{Improving Table Representations.} When passing the previous state in the context, we simply linearise the table. That is, we represent the previous states as discrete tokens passed in the input context for the next turn. This allowed us to use the T5 architecture without modification. A promising direction  for future work is to use continuous representations for the state table~\citep{wu-etal-2022-memformer}. This representation can potentially require fewer or no tokens to represent the state, thus further improving the efficiency of our approach.

\section{Conclusions}

In this paper, we introduce a novel efficient formulation of the DST task and a new system, \name, specifically designed to enhance efficiency and optimise the capabilities of state-of-the-art sequence-to-sequence models. \name represents the dialogue state as an implicit table and updates it using a sequence-to-sequence model to generate table operations at each turn. Our task formalisation provides a significant efficiency gain (up to \qty{2.4}{\x} speed-up in inference time) compared to cumulative state approaches adopted by the current state-of-the-art DST systems. Moreover, this sizeable improvement comes with a minimal efficiency-accuracy trade-off. In fact, \name outperforms other efficient DST approaches in the literature by more than \integer{3} absolute JGA points on MultiWoz 2.4 and shows a competitive performance with respect to current DST state-of-the-art systems.
\name comes with other advantages: it is simple and general (it makes no assumptions about the schema and does not require any specialised architecture) and it is robust to noise. Moreover, it allows to plug and play sequence-to-sequence models without any architectural modification easily.  Finally, our approach goes beyond the dialogue setting and can be adapted to the sequential processing of long documents for information extraction tasks with memory-constrained language models.

\section*{Acknowledgements}
We thank the anonymous reviewers for their helpful questions and comments, which have helped us improve the quality of the paper.
We sincerely thank Miguel Ballesteros, Yogarshi Vyas, Neha Anna John, Yi Zhang, Paula Czarnowska, Laura Aina, Thomas Mueller, and Marcello Federico for their constructive and detailed feedback on the early versions of the paper. We thank Tiago Pimentel, Josef Valvoda, Davide Lesci, and Marco Lesci for their feedback on the final version.

\section*{Limitations}
In Section~\ref{sec:performance_gap}, we already discussed the limitations and challenges of the model proposed (e.g.,~the model has access to less contextual information from the conversation history, errors can propagate more easily as it does not re-predict the entire cumulative state at each step, and mistakes could only be fixed by explicit delete or update operation). In the following, we concentrate on the limitations that refer to the scope of this work.

\paragraph{Languages.} We experimented with a limited number of languages (English) and datasets (MultiWoz 2.1, 2.2 and 2.4). We do not have experimental evidence that our method can work for other languages, including languages with a richer morphology. Still, our system has been built without any language-specific constraints or resources, other than the T5 checkpoints and the manually annotated training set. Our method can be applied to any other language {(without modification)} for which these resources are available, or by applying cross-lingual techniques and resources (e.g., multilingual language models, translation/projection of the training set) to transfer to other languages zero-shot. In those cases, the expected quality is lower, but the efficiency advantage of \name remains. 

\paragraph{Models.} We experimented with two models (T5v1.1 base and large). This is due to the restriction on our computational budget to be both economically- and environmentally-friendly, which made it infeasible to conduct thorough experiments using larger-scale language models. However, we re-emphasise that \name allows one to easily plug and play arbitrary language models and the efficiency advantage of \name remains.

\paragraph{Diversity in the Evaluation Dataset.} We experimented with three different versions of the MultiWoz dataset (2.1, 2.2, and 2.4). Although this is the current benchmark for DST accepted by the community, and we followed the standard evaluation methodology and metrics, we are aware that the results presented might not be directly generalisable to other datasets or real-world scenarios with a considerable data shift with respect to MultiWoz. Additionally, MultiWoz has a certain level of noise and this can have an impact on the evaluation and the generalisation capabilities of the model trained.

\bibliography{assets/biblio}
\bibliographystyle{assets/acl_natbib}

\clearpage

\appendix

\section*{Appendix for \enquote{\emph{Diable}: Efficient \underline{Dia}logue State Tracking as Operations on Ta\underline{ble}s}}

\section{Data Statistics}\label{sec:data_stats}

In this section, we report statistics about versions 2.1-2.4 of the MultiWoz dataset. \Cref{tab:domain_stats} shows the distribution of domains across dialogues and turns. \Cref{tab:slot_stats} reports the distribution of slots across dialogues and turns. Finally, \Cref{tab:general_stats} reports general statistics regarding the frequency of domains, slots, and turns.

\section{Data Preprocessing}\label{appendix:data_preprocessing}

\paragraph{Data Cleaning.} Following the MultiWoz convention, we filter out dialogues in the \enquote{bus}, \enquote{police} and \enquote{hospital} domains (and the respective slots from multi-domain dialogues), and we remove the invalid dialogue \code{SNG01862.json}. The processed dataset contains 5 domains (\enquote{restaurant}, \enquote{train}, \enquote{hotel}, \enquote{taxi}, \enquote{attraction}), \integer{30} slots, \integer{9917} dialogues, and \integer{79793} turns.

\paragraph{Slot-Value Representation.} We represent states as a list of triplets in the form of \code{(domain, slot, value)}. We define a slot as the concatenation of the domain name and slot name, e.g.,~\code{(restaurant-area, center)}. Annotations can possibly contain multiple acceptable values. During testing, this is not problematic as we consider a prediction correct when the predicted value is contained in the acceptable set of values. However, during training, we need to choose one in order to use teacher forcing. To do so, we first check which one of the possible values is actually a span from the text. If none are present, we choose the longest. Since the casing is inconsistent, we lowercase all the values.

\paragraph{Label Normalisation.} MultiWoz contains noisy annotations and different authors have tried to alleviate the issue by devising different label normalisation procedures. For example, the scripts by \citet{wu-etal-2019-transferable}\footnote{\url{https://github.com/jasonwu0731/trade-dst/blob/master/utils/fix_label.py}.} and \citet{hosseini-als-etal-2020-simple}\footnote{\url{https://github.com/salesforce/simpletod/tree/master/noisy_annotations}.}. In this work, we try to balance being as faithful as possible to the original annotations without needlessly penalizing the evaluation of our system. In detail, we target the following noisy annotation values:
\begin{itemize}
    \item Typos: \enquote{guest house}, \enquote{swimming pool}, \enquote{night club}, \enquote{concert hall} that appear with and without spaces. When one is present, we also add the other version as a possible correct answer. This normalisation affects \enquote{hotel-type}, \enquote{attraction-type}, \enquote{hotel-name}, and \enquote{attraction-name} slots.
    \item Spelling inconsistencies: \enquote{theater} and \enquote{center} appear both in their UK and US English versions. When one is present, we add the alternative version. This normalisation affects the \enquote{hotel-area}, \enquote{restaurant-area}, \enquote{attraction-area}, and \enquote{attraction-type}.
    \item Names starting with \enquote{the}: some names appear with the \enquote{the} proposition. In such cases, we add a version without the proposition. This normalisation affects the \enquote{hotel-name},\enquote{restaurant-name}, and \enquote{attraction-name} slots.
    \item Categorical slots: the \enquote{hotel-star} slot is a categorical slot whose values are integers in 0-8. In some cases, the annotation includes the literal \enquote{star} string. In such cases, we remove the \enquote{star} from the annotation.
\end{itemize}
Overall, these are minimal changes. Many such errors caused by noisy labels are still present in the dataset. We leave as future work the creation of an even cleaner evaluation dataset. More details on the impact of these normalisations are available in \Cref{appendix:all_results}.

\begin{table}[!t]
    \centering
    \small
    \begin{adjustbox}{width=\columnwidth}
    \begin{tabular}{lrrrr}
\toprule
\textbf{Domain} & \textbf{2.1} & \textbf{2.2} & \textbf{2.3} & \textbf{2.4} \\
\midrule 
\midrule 
\textbf{Number of dialogues} \\
\midrule 
attraction &   3494 &   3484 &   3503 &   3486 \\
hotel &   4190 &   4182 &   4228 &   4188 \\
restaurant &   4748 &   4728 &   4765 &   4732 \\
taxi &   1879 &   1872 &   1884 &   1875 \\
train &   3940 &   3931 &   3945 &   3936 \\
\midrule 
\midrule 
\textbf{Number of turns} \\
\midrule 
attraction &  24016 &  23940 &  24144 &  23986 \\
 hotel &  31416 &  31378 &  31735 &  31399 \\
 restaurant &  33201 &  33104 &  33369 &  33121 \\
 taxi &   7760 &   7708 &   7833 &   7739 \\
 train &  27738 &  27699 &  27830 &  27721 \\
\bottomrule
\end{tabular}
    \end{adjustbox}
    \caption{Frequency of domains across dialogues and turns for MultiWoz 2.1-2.4.}
    \label{tab:domain_stats}
\end{table}

\paragraph{Input Creation.} In a preliminary study, we experimented with different possible templates for the input sequence and found that, after a certain degree, adding more text to the prompt was not beneficial and the exact wording was not having a big impact. Therefore, to balance simplicity and accuracy, for all experiments, we used the following simple templates,
\begin{itemize}
\itemsep0em
    \item \textit{light}Cumulative: 
    \begin{tcolorbox}
    \code{generate full state: dialogue: system: \{system\_utterance\} user: \{user\_utterance\} <sep> previous dialogue states: \{previous\_states\} <sep> history: \{history\} <sep> states:}
    \end{tcolorbox}
    \item \name 
    \begin{tcolorbox}
    \code{generate update operations: dialogue: system: \{system\_utterance\} user: \{user\_utterance\} <sep> previous dialogue states: \{previous\_states\} <sep> history: \{history\} <sep> operations:}
    \end{tcolorbox}
\end{itemize}
In our experiments, up to decimal differences, including the schema in the input context does not affect performance. Similarly, there is no impact from excluding the \code{<sep>} token to separate the various parts of the input context. However, excluding the prefixes---i.e., \code{generate update operations} and \code{generate full state}---reduces performances by up to \qty{1.2}{\percent} JGA for both state operations and cumulative state models. A similar effect is caused by removing the \enquote{system}/\enquote{user} identifiers (as also observed by \citep{hosseini-als-etal-2020-simple}. 

Note that we did not optimise over the choice of the prefixes; given the new instruction fine-tuned models recently proposed \citep{flan-t5}, we hypothesise---and leave for future work---that different prompts can improve the DST, especially in the few-shot setting. Finally, lower-casing the text decreases performances by up to \qty{1}{\percent} JGA. 

An example of an actual processed conversation is shown in \Cref{tab:example}.

\begin{table*}[!ht]
\centering
\small
\begin{adjustbox}{width=\textwidth}
\begin{tabular}{cp{2cm}p{2cm}>{\raggedright\arraybackslash}p{3.5cm}|p{5cm}>{\raggedright\arraybackslash}p{2cm}}
\toprule
    \textbf{Turn} & \multicolumn{2}{c}{\textbf{Utterances}} & \multicolumn{1}{c}{\textbf{State} }& \multicolumn{2}{c}{\textbf{Processed data}} \\
    \cmidrule(lr){2-3}\cmidrule(lr){5-6}
     & \multicolumn{1}{c}{\textbf{System}} & \multicolumn{1}{c}{\textbf{User}} &  & \multicolumn{1}{c}{\textbf{Input}} & \multicolumn{1}{c}{\textbf{Target}} \\
\midrule
    0 &   none & I'm looking for a place to stay. It needs to be a guesthouse and include free wifi. & \{'hotel-internet': 'yes', 'hotel-type': 'guesthouse'\} & generate update operations: dialogue: system: none user: I'm looking for a place to stay. It needs to be a guesthouse and include free wifi. <sep> \textcolor{red}{previous dialogue states: none} <sep> operations: & INSERT hotel-internet = yes ; INSERT hotel-type = guesthouse \\
    \midrule
    1 & There are 23 hotels that meet your needs. Would you like to narrow your search by area and/or price range? & I would like for it to be cheap and include free parking. & \{'hotel-internet': 'yes', 'hotel-parking': 'yes', 'hotel-pricerange': 'cheap', 'hotel-type': 'guesthouse'\} & generate update operations: dialogue: system: There are 23 hotels that meet your needs. Would you like to narrow your search by area and/or price range? user: I would like for it to be cheap and include free parking. <sep> \textcolor{red}{previous dialogue states: hotel-type = guesthouse; hotel-internet = yes} <sep> operations: & INSERT hotel-parking = yes ; INSERT hotel-pricerange = cheap \\
\bottomrule
\end{tabular}
\end{adjustbox}
\caption{Example of a conversation from the MultiWoz dataset (dialogue \code{MUL0003.json}) processed according to our task formalization. The ``Input'' column shows the template used to construct the input sequence. It includes optional fields separated by the \code{<sep>} token. For example, if the context also includes the dialogue history, we add \code{<sep> history: \{history\}} before \code{operations}. In \textcolor{red}{red} the states from the previous turn. We use the value \texttt{none} to fill-in empty utterances (e.g., the first system utterance), states (e.g., the first state is always empty), or no-operations (i.e., when the state does not need to be updated.}
\label{tab:example}
\end{table*}

\section{Training Details}\label{appendix:training_details}

\paragraph{Hardware Details.} We used a server with \integer{8} Tesla V100 GPUs. The batch size on each GPU was limited to \integer{8}. Thus, we ran the majority of the experiments in a multi-GPU setting with \integer{4} GPUs allocated to each training job with no need for gradient accumulation. \name models were trained in \integer{3}-\integer{4} hours, while cumulative state models required \integer{5}-\integer{6} hours. 

Below, we report the output of the \code{lscpu} command (excluding the flags):
\begin{tcolorbox}
\small
\begin{verbatim}
Architecture:        x86_64
CPU op-mode(s):      32-bit, 64-bit
Byte Order:          Little Endian
CPU(s):              64
On-line CPU(s) list: 0-63
Thread(s) per core:  2
Core(s) per socket:  16
Socket(s):           2
NUMA node(s):        2
Vendor ID:           GenuineIntel
CPU family:          6
Model:               79
Model name:          Intel(R) Xeon(R)
                      CPU E5-2686 v4
                      @ 2.30GHz
Stepping:            1
CPU MHz:             2700.216
CPU max MHz:         3000.0000
CPU min MHz:         1200.0000
BogoMIPS:            4600.04
Hypervisor vendor:   Xen
Virtualization type: full
L1d cache:           32K
L1i cache:           32K
L2 cache:            256K
L3 cache:            46080K
NUMA node0 CPU(s):   0-15,32-47
NUMA node1 CPU(s):   16-31,48-63
\end{verbatim}
\end{tcolorbox}

\paragraph{Model.} For all experiments, we use the T5 architecture \citep{raffel-etal-2020-t5} and use the associated T5v1.1 base\footnote{\url{https://huggingface.co/google/t5-v1_1-base}.} and large\footnote{\url{https://huggingface.co/google/t5-v1_1-large}.} checkpoints available on the HuggingFace Hub via the transformers \citep{wolf-etal-2020-transformers} library and implemented using the PyTorch \citep{pytorch} framework. In a preliminary study, we compared T5v1.1 with the original T5 and flan-T5 \citep{flan-t5} variants and did not see any significant differences; we choose to use the T5v1.1 checkpoint since it is not fine-tuned on downstream tasks.

\paragraph{Data Preparation.} We use the default SentencePiece \citep{kudo-richardson-2018-sentencepiece} tokenizer with vocabulary size 32k associated with the T5v1.1 checkpoint and available in the tokenizers library \citep{wolf-etal-2020-transformers}. We add \code{<sep>} to the vocabulary as a special separator token. We truncate only the input sequence at \integer{512} tokens during training but do not truncate during evaluation in order to not penalize cumulative state models (our main baseline).

\paragraph{Training.} We use the Pytorch-Lightning\footnote{\url{https://github.com/Lightning-AI/lightning/}.} library to implement the training loop. For all experiments, we use the Adafactor optimiser \citep{pmlr-v80-shazeer18a} with no weight decay, \code{eps=[1e-30, 0.001]}, \code{clip\_threshold: 1.0}, \code{decay\_rate: -0.8}, \code{beta1: null}, \code{scale\_parameter: false}, \code{relative\_step: false}, and \code{warmup\_init: false}. We set the learning rate to $10^{-4}$ and use a constant schedule. We fix the training budget at \qty{40}{\thousands} optimiser steps and set the batch size to \integer{32} to trade off the speed and precision of the gradient estimation. 

\paragraph{Inference.} For all experiments, we use beam search decoding, as implemented in the HuggingFace library \citep{wolf-etal-2020-transformers} with \integer{4} beams and no additional settings.

\paragraph{Reproducibility.}\label{appendix:reproducibility_hardware} For reproducibility, we use the pseudo-random seed for both data shuffling and model initialization. In each experiment, we fix the seed for pseudo-random number generators and use CUDA deterministic operations. In particular, we use the \code{seed\_everything} function from Pytorch-Lightning\footnote{\url{https://github.com/Lightning-AI/lightning/blob/94e6d52b7e2f2a9ffc21f7e11e087808666fe710/src/lightning_lite/utilities/seed.py\#L20}} to set the seed for pseudo-random number generators in \code{pytorch}, \code{numpy}, and \code{python.random}. In addition, it sets the following environment variables: \code{PL\_GLOBAL\_SEED} that is passed to spawned subprocesses (e.g. ddp\_spawn backend) and \code{PL\_SEED\_WORKERS}. 

\paragraph{Hyper-parameters.} In our initial exploration, we used the default hyper-parameters suggested in the T5 paper \citep{raffel-etal-2020-t5} and on the HuggingFace blog\footnote{\url{https://discuss.huggingface.co/t/t5-finetuning-tips/684}.}, that is batch size \integer{128} and constant learning rate equal to $10^{-3}$. Given the size of MultiWoz, it roughly corresponded to \qty{4}{\thousands} update steps. However, this budget proved to be insufficient. In particular, our own re-implementation of the cumulative state type of models was not in line with the results reported in the literature. More importantly, our \name model was clearly undertrained as demonstrated by the fact that model selection on the validation set was consistently selecting the last checkpoint. Therefore, we scaled up the training budget by \qty{10}{\x} to roughly \qty{40}{\thousands} update steps. We rescaled the batch size to \integer{32} and, consistently, the learning rate to $10^{-4}$---a similar setup is used by \citet{zhao-etal-2022-description}. This new training budget corresponds to roughly \integer{20} epochs. We did not notice any significant improvement by further increasing it. Finally, we experimented with both Adafactor and AdamW \citep{adamw} and the former consistently outperformed the latter while also speeding up the training process.

\section{Complete Tables of Results}\label{appendix:all_results}

In this section, we report the complete set of results for our \name system and \textit{light}Cumulative (our own reproduction of a cumulative state model). We run each experiment with 5 different random seeds and report statistics across runs. Furthermore, we show the effect of label normalisation in rows identified by \enquote{fix label}.

\Cref{tab:large} shows the JGA on the evaluation sets of MultiWoz 2.1-2.4 for the T5v1.1-large models trained on the MultiWoz 2.2. \Cref{tab:base_21} reports the JGA on the evaluation sets of MultiWoz 2.1-2.4 for T5v1.1-base models trained on the MultiWoz 2.1. Finally, \Cref{tab:base_22} contains the JGA on the evaluation sets of MultiWoz 2.1-2.4 for T5v1.1-base models trained on the MultiWoz 2.2.

\begin{table}[!t]
    \centering
    \small
    \begin{adjustbox}{width=\columnwidth}
    \begin{tabular}{llr}
\toprule
\textbf{Contex} &  \textbf{Dataset version} &  JGA \\
\midrule
\midrule
\textbf{Cumulative state} \\
\midrule
$D_{1:t}$ &     2.1 & 51.06 \\
$D_{1:t}$ &     2.2 & 57.30 \\
$D_{1:t}$ &     2.3 & 48.21 \\
$D_{1:t}$ &     2.4 & 58.52 \\
$D_{1:t}$ &     2.1 (fix labels) & 51.94 \\
$D_{1:t}$ &     2.2 (fix labels) & 57.37 \\
$D_{1:t}$ &     2.3 (fix labels) & 49.77 \\
$D_{1:t}$ &     2.4 (fix labels) & 65.82 \\
\midrule
\midrule
\textbf{\emph{Diable}} \\
\midrule
$D_{t-4:t} + \mathcal{B}_{t-1}$ &     2.1 & 49.86 \\
$D_{t-4:t} + \mathcal{B}_{t-1}$ &     2.2 & 56.42 \\
$D_{t-4:t} + \mathcal{B}_{t-1}$ &     2.3 & 47.84 \\
$D_{t-4:t} + \mathcal{B}_{t-1}$ &     2.4 & 58.00 \\
$D_{t-4:t} + \mathcal{B}_{t-1}$ & 2.1 (fix labels) & 50.92 \\
$D_{t-4:t} + \mathcal{B}_{t-1}$ & 2.2 (fix labels) & 56.74 \\
$D_{t-4:t} + \mathcal{B}_{t-1}$ & 2.3 (fix labels) & 49.70 \\
$D_{t-4:t} + \mathcal{B}_{t-1}$ & 2.4 (fix labels) & 65.01 \\
\bottomrule
\end{tabular}
    \end{adjustbox}
    \caption{JGA on the evaluation sets of MultiWoz 2.1-2.4 for T5v1.1-large models trained on the MultiWoz 2.2 training set. Result statistics obtained across \integer{5} random seeds. The evaluation also includes the raw metrics with no label normalisation.}
    \label{tab:large}
\end{table}

\onecolumn

\begin{table*}[!h]
    \small
    \centering
    \begin{adjustbox}{max totalheight=0.45\textheight}
    \begin{tabular}{lrrrrrrrr}
    \toprule
    \textbf{Slots} & \multicolumn{4}{c}{\textbf{Number of dialogues}} & \multicolumn{4}{c}{\textbf{Number of turns}} \\
    \cmidrule(lr){2-5}\cmidrule(lr){6-9}
     &          \textbf{2.1} &    \textbf{2.2} &    \textbf{2.3} &    \textbf{2.4} &      \textbf{2.1} &     \textbf{2.2} &     \textbf{2.3} &     \textbf{2.4} \\
    \midrule
    attraction-area       &        2397 &  2396 &  2431 &  2396 &   16232 &  16401 &  16521 &  16277 \\
    attraction-name       &        2270 &  2260 &  3348 &  2358 &   12557 &  12710 &  18348 &  13041 \\
    attraction-type       &        2502 &  2500 &  2542 &  2503 &   16900 &  17003 &  17203 &  16940 \\
    \midrule
    hotel-area            &        2416 &  2417 &  2478 &  2452 &   17213 &  17562 &  17457 &  17490 \\
    hotel-day             &        2599 &  2599 &  2620 &  2597 &   13927 &  13963 &  14141 &  13928 \\
    hotel-internet        &        1772 &  1772 &  1953 &  1786 &   12706 &  12920 &  13395 &  12814 \\
    hotel-name            &        3039 &  3542 &  3787 &  3154 &   16770 &  19789 &  21920 &  17431 \\
    hotel-parking         &        1819 &  1819 &  2001 &  1848 &   12903 &  13107 &  13617 &  13095 \\
    hotel-people          &        2605 &  2605 &  2631 &  2606 &   13889 &  13952 &  14123 &  13935 \\
    hotel-pricerange      &        2244 &  2246 &  2324 &  2251 &   16070 &  16337 &  16327 &  16146 \\
    hotel-stars           &        1984 &  1984 &  2035 &  1969 &   14220 &  14475 &  14435 &  14177 \\
    hotel-stay            &        2605 &  2605 &  2624 &  2605 &   13960 &  14024 &  14168 &  13993 \\
    hotel-type            &        2262 &  2263 &  2728 &  2242 &   16339 &  16733 &  17891 &  16344 \\
    \midrule
    restaurant-area       &        3414 &  3412 &  3461 &  3400 &   22915 &  23237 &  23119 &  22894 \\
    restaurant-day        &        2626 &  2626 &  2643 &  2626 &   14463 &  14485 &  14643 &  14492 \\
    restaurant-food       &        3605 &  3599 &  3647 &  3605 &   24545 &  24841 &  24822 &  24594 \\
    restaurant-name       &        3230 &  3850 &  4298 &  3358 &   16946 &  20249 &  22978 &  17690 \\
    restaurant-people     &        2635 &  2635 &  2655 &  2638 &   14543 &  14590 &  14747 &  14591 \\
    restaurant-pricerange &        3331 &  3330 &  3381 &  3314 &   22498 &  22759 &  22793 &  22472 \\
    restaurant-time       &        2620 &  2616 &  2649 &  2621 &   14346 &  14410 &  14548 &  14374 \\
    \midrule
    taxi-arriveby         &         846 &   840 &   861 &   845 &    3122 &   3158 &   3190 &   3134 \\
    taxi-departure        &        1836 &  1826 &  1840 &  1832 &    7089 &   7087 &   7124 &   7077 \\
    taxi-destination      &        1837 &  1831 &  1841 &  1829 &    7091 &   7117 &   7149 &   7073 \\
    taxi-leaveat          &        1071 &  1051 &  1094 &  1063 &    3963 &   3929 &   4077 &   3934 \\
    \midrule
    train-arriveby        &        2111 &  2106 &  2136 &  2106 &   13097 &  13149 &  13220 &  13086 \\
    train-day             &        3793 &  3792 &  3828 &  3793 &   24656 &  24680 &  24854 &  24678 \\
    train-departure       &        3783 &  3774 &  3832 &  3768 &   24782 &  24792 &  25185 &  24704 \\
    train-destination     &        3806 &  3799 &  3831 &  3796 &   25254 &  25247 &  25458 &  25201 \\
    train-leaveat         &        2035 &  2023 &  2085 &  2038 &   12622 &  12838 &  12802 &  12753 \\
    train-people          &        2266 &  2266 &  2284 &  2225 &   10858 &  10932 &  11043 &  10700 \\
    \bottomrule
    \end{tabular}%
    \end{adjustbox}
    \caption{Frequency of slots across dialogues and turns for MultiWoz 2.1-2.4.}
    \label{tab:slot_stats}
\end{table*}

\begin{table*}[!ht]
    \centering
    \small
    \begin{adjustbox}{max totalheight=0.4\textheight}
\begin{tabular}{lrrrrrrrrrrrr}
\toprule
 & \multicolumn{4}{c}{\textbf{Train}} & \multicolumn{4}{c}{\textbf{Validation}} & \multicolumn{4}{c}{\textbf{Test}} \\
         \cmidrule(lr){2-5}\cmidrule(lr){6-9}\cmidrule(lr){10-13}
        \textbf{Version} & \textbf{2.1} &\textbf{2.2} &\textbf{2.3} &\textbf{2.4} &  \textbf{2.1} &\textbf{2.2} &\textbf{2.3} &\textbf{2.4} &\textbf{2.1} &\textbf{2.2} &\textbf{2.3} &\textbf{2.4} \\
\midrule
\midrule
\textbf{Number of domains} \\
\midrule
        mean &     1.81 &     1.81 &     1.82 &     1.81 &       1.97 &     1.96 &     1.97 &     1.95 &     1.93 &     1.93 &     1.94 &     1.92 \\
        std &     0.69 &     0.68 &     0.69 &     0.69 &       0.62 &     0.62 &     0.63 &     0.61 &     0.63 &     0.62 &     0.63 &     0.62 \\
        min &     1 &     1 &     1 &     1 &       1 &     1 &     1 &     1 &     1 &     1 &     1 &     1 \\
         25\% &     1 &     1 &     1 &     1 &       2 &     2 &     2 &     2 &     2 &     2 &     2 &     2 \\
         50\% &     2 &     2 &     2 &     2 &       2 &     2 &     2 &     2 &     2 &     2 &     2 &     2 \\
         75\% &     2 &     2 &     2 &     2 &       2 &     2 &     2 &     2 &     2 &     2 &     2 &     2 \\
         max &     5 &     4 &     5 &     5 &       4 &     4 &     4 &     4 &     4 &     4 &     4 &     4 \\
        \midrule
        \midrule
\textbf{Number of turns} \\
        \midrule
         mean &     7.96 &     7.96 &     7.96 &     7.96 &       8.37 &     8.37 &     8.37 &     8.37 &     8.37 &     8.37 &     8.37 &     8.37 \\
         std &     2.59 &     2.59 &     2.59 &     2.59 &       2.24 &     2.24 &     2.24 &     2.24 &     2.37 &     2.37 &     2.37 &     2.37 \\
         min &     2 &     2 &     2 &     2 &       3 &     3 &     3 &     3 &     3 &     3 &     3 &     3 \\
         25\% &     6 &     6 &     6 &     6 &       7 &     7 &     7 &     7 &     7 &     7 &     7 &     7 \\
         50\% &     8 &     8 &     8 &     8 &       8 &     8 &     8 &     8 &     8 &     8 &     8 &     8 \\
         75\% &    10 &    10 &    10 &    10 &      10 &    10 &    10 &    10 &    10 &    10 &    10 &    10 \\
         max &    23 &    23 &    23 &    23 &      18 &    18 &    18 &    18 &    19 &    19 &    19 &    19 \\
        \midrule
        \midrule
\textbf{Number of slots} \\
        \midrule    
         mean &     7.49 &     7.59 &     7.95 &     7.49 &       7.97 &     8.10 &     8.44 &     8.23 &     8.08 &     8.16 &     8.45 &     8.08 \\
         std &     3.51 &     3.59 &     3.62 &     3.51 &       3.13 &     3.23 &     3.23 &     3.26 &     3.20 &     3.27 &     3.27 &     3.20 \\
         min &     1 &     1 &     1 &     1 &       1 &     1 &     1 &     1 &     1 &     1 &     1 &     1 \\
         25\% &     5 &     5 &     5 &     5 &       6 &     6 &     6 &     6 &     6 &     6 &     6 &     6 \\
         50\% &     7 &     7 &     8 &     7 &       8 &     8 &     8 &     8 &     8 &     8 &     8 &     8 \\
         75\% &    10 &    10 &    11 &    10 &      10 &    11 &    11 &    11 &    10 &    11 &    11 &    10 \\
         max &    20 &    20 &    20 &    20 &      17 &    18 &    17 &    17 &    18 &    19 &    18 &    19 \\
\bottomrule
\end{tabular}
    \end{adjustbox}
    \caption{Number of domains, turns, and slots per dialogue for MultiWoz 2.1-2.4.}
    \label{tab:general_stats}
\end{table*}

\begin{table}[!t]
\small
\centering
%\begin{adjustbox}{max totalheight=0.8\textheight}
\begin{adjustbox}{width=0.8\textwidth}
\begin{tabular}{llrrrrrrr}
\toprule
        \textbf{Context} &  \textbf{Dataset version} &  \textbf{mean} &  \textbf{std} &   \textbf{min} &   \textbf{25\%} &   \textbf{50\%} &   \textbf{75\%} &   \textbf{max} \\
\midrule
\midrule
\textbf{Cumulative state} \\
\midrule
   $D_{1:t}$ &              2.1 & 53.80 & 0.65 & 52.81 & 53.57 & 54.02 & 54.10 & 54.53 \\
   $D_{1:t}$ &              2.2 & 53.48 & 0.58 & 52.65 & 53.30 & 53.49 & 53.74 & 54.22 \\
   $D_{1:t}$ &              2.3 & 49.33 & 0.59 & 48.43 & 49.29 & 49.31 & 49.62 & 50.01 \\
   $D_{1:t}$ &              2.4 & 60.54 & 0.57 & 60.01 & 60.22 & 60.42 & 60.74 & 61.33 \\
   $D_{1:t}$ & 2.1 (fix labels) & 53.91 & 0.63 & 52.96 & 53.74 & 54.07 & 54.11 & 54.67 \\
   $D_{1:t}$ & 2.2 (fix labels) & 54.60 & 0.61 & 53.77 & 54.37 & 54.46 & 55.17 & 55.25 \\
   $D_{1:t}$ & 2.3 (fix labels) & 51.16 & 0.40 & 50.66 & 50.84 & 51.29 & 51.38 & 51.63 \\
   $D_{1:t}$ & 2.4 (fix labels) & 67.56 & 0.52 & 67.13 & 67.27 & 67.34 & 67.65 & 68.43 \\
   \midrule
\midrule
\textbf{\emph{Diable}} \\
\midrule
$D_{t-4:t} + \mathcal{B}_{t-1}$ &              2.1 & 53.80 & 0.65 & 52.75 & 53.78 & 53.84 & 54.11 & 54.50 \\
$D_{t-4:t} + \mathcal{B}_{t-1}$ &              2.2 & 53.86 & 0.41 & 53.31 & 53.65 & 53.87 & 54.11 & 54.38 \\
$D_{t-4:t} + \mathcal{B}_{t-1}$ &              2.3 & 49.70 & 0.57 & 48.96 & 49.38 & 49.76 & 49.93 & 50.46 \\
$D_{t-4:t} + \mathcal{B}_{t-1}$ &              2.4 & 62.80 & 0.99 & 61.06 & 63.13 & 63.14 & 63.19 & 63.48 \\
$D_{t-4:t} + \mathcal{B}_{t-1}$ & 2.1 (fix labels) & 53.97 & 0.66 & 52.90 & 53.92 & 54.07 & 54.33 & 54.65 \\
$D_{t-4:t} + \mathcal{B}_{t-1}$ & 2.2 (fix labels) & 54.58 & 0.67 & 53.51 & 54.48 & 54.63 & 55.02 & 55.25 \\
$D_{t-4:t} + \mathcal{B}_{t-1}$ & 2.3 (fix labels) & 51.65 & 0.46 & 51.04 & 51.33 & 51.76 & 52.06 & 52.08 \\
$D_{t-4:t} + \mathcal{B}_{t-1}$ & 2.4 (fix labels) & 70.46 & 1.18 & 68.49 & 70.60 & 70.66 & 70.90 & 71.64 \\
          $\mathcal{B}_{t-1}$ &              2.1 & 53.58 & 0.30 & 53.16 & 53.46 & 53.59 & 53.73 & 53.96 \\
          $\mathcal{B}_{t-1}$ &              2.2 & 53.03 & 0.23 & 52.78 & 52.81 & 53.08 & 53.20 & 53.30 \\
          $\mathcal{B}_{t-1}$ &              2.3 & 49.30 & 0.85 & 48.43 & 48.82 & 49.12 & 49.50 & 50.65 \\
          $\mathcal{B}_{t-1}$ &              2.4 & 61.48 & 0.83 & 60.85 & 61.01 & 61.26 & 61.33 & 62.93 \\
          $\mathcal{B}_{t-1}$ & 2.1 (fix labels) & 53.91 & 0.70 & 53.31 & 53.49 & 53.72 & 53.99 & 55.07 \\
          $\mathcal{B}_{t-1}$ & 2.2 (fix labels) & 54.41 & 0.93 & 53.73 & 53.91 & 54.02 & 54.35 & 56.02 \\
          $\mathcal{B}_{t-1}$ & 2.3 (fix labels) & 51.48 & 0.70 & 51.07 & 51.13 & 51.17 & 51.29 & 52.73 \\
          $\mathcal{B}_{t-1}$ & 2.4 (fix labels) & 70.03 & 0.95 & 68.79 & 69.25 & 69.76 & 70.55 & 71.42 \\
\bottomrule
\end{tabular}

\end{adjustbox}
    \caption{JGA on the test sets of MultiWoz 2.1-2.4 for T5v1.1-base models trained on the MultiWoz 2.1 training set. Result statistics obtained across \integer{5} random seeds. The evaluation also includes the raw metrics with no label normalisation.}
    \label{tab:base_21}
\end{table}

\begin{table}[!t]
\small
\centering
\begin{adjustbox}{max totalheight=0.95\textheight}
\begin{tabular}{llrrrrrrr}
\toprule
\textbf{Context} & \textbf{ Dataset version} &  \textbf{mean} &  \textbf{std} &   \textbf{min} &   \textbf{25\%} &   \textbf{50\%} &   \textbf{75\%} &   \textbf{max} \\
\midrule
\midrule
\multicolumn{2}{l}{\textbf{Cumulative state}} & & & & & & & \\
\midrule
        $D_{1:t}$ &              2.1 & 49.47 & 0.22 & 49.12 & 49.39 & 49.55 & 49.63 & 49.66 \\
        $D_{1:t}$ &              2.2 & 56.72 & 0.49 & 56.17 & 56.38 & 56.61 & 57.18 & 57.28 \\
        $D_{1:t}$ &              2.3 & 47.67 & 0.59 & 47.10 & 47.19 & 47.48 & 48.14 & 48.44 \\
        $D_{1:t}$ &              2.4 & 56.89 & 0.51 & 56.16 & 56.65 & 57.04 & 57.08 & 57.50 \\
        $D_{1:t}$ & 2.1 (fix labels) & 50.69 & 0.37 & 50.45 & 50.47 & 50.60 & 50.61 & 51.34 \\
        $D_{1:t}$ & 2.2 (fix labels) & 57.01 & 0.45 & 56.43 & 56.81 & 56.89 & 57.42 & 57.51 \\
        $D_{1:t}$ & 2.3 (fix labels) & 49.51 & 0.63 & 48.74 & 48.94 & 49.81 & 49.85 & 50.19 \\
        $D_{1:t}$ & 2.4 (fix labels) & 63.11 & 0.83 & 61.98 & 62.89 & 63.13 & 63.27 & 64.30 \\
$D_{1:t} + \mathcal{B}_{t-1}$ &              2.1 & 49.50 & 0.50 & 48.91 & 49.12 & 49.59 & 49.72 & 50.18 \\
$D_{1:t} + \mathcal{B}_{t-1}$ &              2.2 & 56.43 & 0.73 & 55.41 & 56.20 & 56.28 & 56.96 & 57.28 \\
$D_{1:t} + \mathcal{B}_{t-1}$ &              2.3 & 47.63 & 0.57 & 46.96 & 47.26 & 47.52 & 48.14 & 48.29 \\
$D_{1:t} + \mathcal{B}_{t-1}$ &              2.4 & 57.54 & 0.55 & 56.76 & 57.15 & 57.84 & 57.95 & 57.99 \\
$D_{1:t} + \mathcal{B}_{t-1}$ & 2.1 (fix labels) & 50.71 & 0.55 & 50.04 & 50.47 & 50.57 & 50.98 & 51.51 \\
$D_{1:t} + \mathcal{B}_{t-1}$ & 2.2 (fix labels) & 56.89 & 0.57 & 56.19 & 56.65 & 56.67 & 57.33 & 57.60 \\
$D_{1:t} + \mathcal{B}_{t-1}$ & 2.3 (fix labels) & 49.45 & 0.54 & 48.81 & 49.08 & 49.50 & 49.67 & 50.20 \\
$D_{1:t} + \mathcal{B}_{t-1}$ & 2.4 (fix labels) & 63.27 & 0.68 & 62.11 & 63.27 & 63.48 & 63.74 & 63.75 \\
               $\mathcal{B}_{t-1}$ &              2.1 & 49.59 & 0.26 & 49.16 & 49.57 & 49.65 & 49.74 & 49.84 \\
               $\mathcal{B}_{t-1}$ &              2.2 & 56.38 & 0.46 & 55.68 & 56.20 & 56.43 & 56.69 & 56.88 \\
               $\mathcal{B}_{t-1}$ &              2.3 & 47.20 & 0.79 & 46.13 & 46.72 & 47.30 & 47.80 & 48.06 \\
               $\mathcal{B}_{t-1}$ &              2.4 & 57.35 & 1.21 & 55.22 & 57.54 & 57.89 & 57.99 & 58.13 \\
               $\mathcal{B}_{t-1}$ & 2.1 (fix labels) & 50.90 & 0.11 & 50.77 & 50.83 & 50.91 & 50.95 & 51.06 \\
               $\mathcal{B}_{t-1}$ & 2.2 (fix labels) & 56.50 & 0.47 & 55.93 & 56.13 & 56.57 & 56.81 & 57.08 \\
               $\mathcal{B}_{t-1}$ & 2.3 (fix labels) & 49.08 & 0.90 & 47.90 & 48.48 & 49.23 & 49.65 & 50.15 \\
               $\mathcal{B}_{t-1}$ & 2.4 (fix labels) & 63.52 & 0.96 & 62.22 & 62.89 & 63.73 & 64.24 & 64.53 \\
     $D_{t-4:t} + \mathcal{B}_{t-1}$ &              2.1 & 49.38 & 0.27 & 49.08 & 49.09 & 49.54 & 49.59 & 49.61 \\
     $D_{t-4:t} + \mathcal{B}_{t-1}$ &              2.2 & 56.82 & 0.92 & 55.48 & 56.25 & 57.23 & 57.49 & 57.64 \\
     $D_{t-4:t} + \mathcal{B}_{t-1}$ &              2.3 & 47.52 & 0.72 & 46.65 & 47.12 & 47.38 & 47.98 & 48.48 \\
     $D_{t-4:t} + \mathcal{B}_{t-1}$ &              2.4 & 57.71 & 0.86 & 56.73 & 57.01 & 57.68 & 58.49 & 58.64 \\
     $D_{t-4:t} + \mathcal{B}_{t-1}$ & 2.1 (fix labels) & 50.58 & 0.34 & 50.11 & 50.34 & 50.76 & 50.77 & 50.90 \\
     $D_{t-4:t} + \mathcal{B}_{t-1}$ & 2.2 (fix labels) & 57.12 & 0.93 & 55.70 & 56.69 & 57.45 & 57.81 & 57.95 \\
     $D_{t-4:t} + \mathcal{B}_{t-1}$ & 2.3 (fix labels) & 49.58 & 0.97 & 48.03 & 49.27 & 50.03 & 50.09 & 50.46 \\
     $D_{t-4:t} + \mathcal{B}_{t-1}$ & 2.4 (fix labels) & 63.15 & 1.56 & 61.58 & 61.96 & 62.70 & 64.23 & 65.26 \\
\midrule
\midrule
\multicolumn{2}{l}{\textbf{\emph{Diable}}} & & & & & & & \\
\midrule
        $D_{1:t}$ &              2.1 & 48.74 & 0.20 & 48.48 & 48.67 & 48.67 & 48.91 & 48.98 \\
        $D_{1:t}$ &              2.2 & 55.33 & 0.70 & 54.40 & 55.11 & 55.28 & 55.58 & 56.31 \\
        $D_{1:t}$ &              2.3 & 47.46 & 0.24 & 47.18 & 47.35 & 47.35 & 47.65 & 47.78 \\
        $D_{1:t}$ &              2.4 & 58.76 & 0.57 & 58.25 & 58.29 & 58.68 & 58.94 & 59.64 \\
        $D_{1:t}$ & 2.1 (fix labels) & 50.00 & 0.14 & 49.81 & 49.88 & 50.08 & 50.08 & 50.14 \\
        $D_{1:t}$ & 2.2 (fix labels) & 55.63 & 0.68 & 54.73 & 55.39 & 55.63 & 55.78 & 56.62 \\
        $D_{1:t}$ & 2.3 (fix labels) & 49.31 & 0.26 & 49.09 & 49.13 & 49.21 & 49.40 & 49.72 \\
        $D_{1:t}$ & 2.4 (fix labels) & 64.95 & 0.55 & 64.05 & 64.79 & 65.29 & 65.30 & 65.33 \\
$D_{1:t} + \mathcal{B}_{t-1}$ &              2.1 & 48.86 & 0.30 & 48.49 & 48.68 & 48.82 & 49.04 & 49.27 \\
$D_{1:t} + \mathcal{B}_{t-1}$ &              2.2 & 56.04 & 0.77 & 55.33 & 55.71 & 55.82 & 56.01 & 57.34 \\
$D_{1:t} + \mathcal{B}_{t-1}$ &              2.3 & 48.53 & 0.66 & 47.82 & 48.17 & 48.40 & 48.71 & 49.57 \\
$D_{1:t} + \mathcal{B}_{t-1}$ &              2.4 & 59.60 & 0.54 & 58.94 & 59.10 & 59.86 & 59.94 & 60.16 \\
$D_{1:t} + \mathcal{B}_{t-1}$ & 2.1 (fix labels) & 50.00 & 0.30 & 49.62 & 49.86 & 49.92 & 50.22 & 50.39 \\
$D_{1:t} + \mathcal{B}_{t-1}$ & 2.2 (fix labels) & 56.28 & 0.84 & 55.44 & 55.87 & 56.19 & 56.24 & 57.66 \\
$D_{1:t} + \mathcal{B}_{t-1}$ & 2.3 (fix labels) & 50.57 & 0.73 & 49.74 & 50.23 & 50.38 & 50.79 & 51.70 \\
$D_{1:t} + \mathcal{B}_{t-1}$ & 2.4 (fix labels) & 66.51 & 0.87 & 65.18 & 66.35 & 66.56 & 66.93 & 67.53 \\
               $\mathcal{B}_{t-1}$ &              2.1 & 49.03 & 0.24 & 48.74 & 48.93 & 48.96 & 49.17 & 49.35 \\
               $\mathcal{B}_{t-1}$ &              2.2 & 55.99 & 0.66 & 55.40 & 55.67 & 55.81 & 55.94 & 57.12 \\
               $\mathcal{B}_{t-1}$ &              2.3 & 47.81 & 0.34 & 47.25 & 47.82 & 47.90 & 47.94 & 48.14 \\
               $\mathcal{B}_{t-1}$ &              2.4 & 59.13 & 0.86 & 58.19 & 58.42 & 59.10 & 59.73 & 60.23 \\
               $\mathcal{B}_{t-1}$ & 2.1 (fix labels) & 50.58 & 0.52 & 50.15 & 50.23 & 50.45 & 50.62 & 51.45 \\
               $\mathcal{B}_{t-1}$ & 2.2 (fix labels) & 56.30 & 0.67 & 55.63 & 55.98 & 56.21 & 56.24 & 57.42 \\
               $\mathcal{B}_{t-1}$ & 2.3 (fix labels) & 49.85 & 0.40 & 49.20 & 49.88 & 49.88 & 49.96 & 50.31 \\
               $\mathcal{B}_{t-1}$ & 2.4 (fix labels) & 66.13 & 0.97 & 65.04 & 65.45 & 66.03 & 66.63 & 67.49 \\
     $D_{t-4:t} + \mathcal{B}_{t-1}$ &              2.1 & 49.27 & 0.42 & 48.82 & 48.87 & 49.31 & 49.67 & 49.69 \\
     $D_{t-4:t} + \mathcal{B}_{t-1}$ &              2.2 & 56.21 & 0.56 & 55.67 & 55.79 & 56.00 & 56.62 & 56.96 \\
     $D_{t-4:t} + \mathcal{B}_{t-1}$ &              2.3 & 48.01 & 0.52 & 47.27 & 47.83 & 47.98 & 48.36 & 48.62 \\
     $D_{t-4:t} + \mathcal{B}_{t-1}$ &              2.4 & 58.86 & 0.31 & 58.53 & 58.68 & 58.76 & 58.98 & 59.32 \\
     $D_{t-4:t} + \mathcal{B}_{t-1}$ & 2.1 (fix labels) & 50.40 & 0.38 & 49.93 & 50.07 & 50.50 & 50.72 & 50.77 \\
     $D_{t-4:t} + \mathcal{B}_{t-1}$ & 2.2 (fix labels) & 56.48 & 0.57 & 55.90 & 56.02 & 56.36 & 56.88 & 57.26 \\
     $D_{t-4:t} + \mathcal{B}_{t-1}$ & 2.3 (fix labels) & 50.00 & 0.63 & 49.04 & 49.86 & 50.05 & 50.31 & 50.73 \\
     $D_{t-4:t} + \mathcal{B}_{t-1}$ & 2.4 (fix labels) & 65.35 & 0.50 & 64.68 & 65.19 & 65.37 & 65.48 & 66.05 \\
     \midrule
     \multicolumn{2}{l}{\textit{Gold previous state}} & & & & & & & \\
     \midrule
$\mathcal{B}_{t-1}$ & 2.1 & 68.44 & 0.14 & 68.20 & 68.41 & 68.49 & 68.50 & 68.58 \\
$\mathcal{B}_{t-1}$ & 2.2 & 82.38 & 0.27 & 81.96 & 82.28 & 82.43 & 82.53 & 82.68 \\
$\mathcal{B}_{t-1}$ & 2.3 & 58.49 & 0.20 & 58.25 & 58.40 & 58.46 & 58.56 & 58.79 \\
$\mathcal{B}_{t-1}$ & 2.4 & 55.38 & 0.25 & 55.20 & 55.20 & 55.21 & 55.60 & 55.71 \\
$\mathcal{B}_{t-1}$ & 2.1 (fix labels) & 70.00 & 0.16 & 69.74 & 69.95 & 70.05 & 70.10 & 70.14 \\
$\mathcal{B}_{t-1}$ & 2.2 (fix labels) & 82.50 & 0.28 & 82.07 & 82.45 & 82.54 & 82.65 & 82.81 \\
$\mathcal{B}_{t-1}$ & 2.3 (fix labels) & 61.06 & 0.22 & 60.78 & 60.95 & 61.04 & 61.19 & 61.34 \\
$\mathcal{B}_{t-1}$ & 2.4 (fix labels) & 61.46 & 0.32 & 61.18 & 61.22 & 61.29 & 61.81 & 61.81 \\
\bottomrule
\end{tabular}

\end{adjustbox}
    \caption{JGA on the test sets of MultiWoz 2.1-2.4 for T5v1.1-base models trained on the MultiWoz 2.2 training set. Result statistics obtained across \integer{5} random seeds. The evaluation also includes the raw metrics with no label normalisation.}
    \label{tab:base_22}
\end{table}

\end{document}